\documentclass[10pt,journal,compsoc]{IEEEtran}
\usepackage{ising_v1}
\usepackage{booktabs}
\usepackage{xcolor}
\usepackage{hyperref}
\usepackage{graphicx}
\graphicspath{{./images}}
\usepackage{amsmath}
\usepackage{multirow}
\usepackage{subcaption}
\usepackage{comment}
\usepackage{wrapfig}
\usepackage{cleveref}
\usepackage{amssymb}
\usepackage{listings}
\usepackage{longtable}
\usepackage{footnote}
\makesavenoteenv{table}
\makesavenoteenv{tabular}
\usepackage{algorithm}
\usepackage{algpseudocode}
\usepackage{threeparttable}
\usepackage{xspace}
\usepackage{wasysym}

\usepackage{listings}
\usepackage{color}

\definecolor{dkgreen}{rgb}{0,0.6,0}
\definecolor{gray}{rgb}{0.5,0.5,0.5}
\definecolor{mauve}{rgb}{0.58,0,0.82}

\newcommand{\fedeval}{FedEval\xspace}
\newtheorem{definition}{Definition}

\def\submission{1}

\newcommand{\highlight}[1]{
\if\submission1
{\color{blue}#1}
\else
#1
\fi
}

\newcommand{\TODO}[1]{
\if\submission1
{\color{red}#1}
\else
#1
\fi
}

\newcommand{\tabincell}[2]{\begin{tabular}{@{}#1@{}}#2\end{tabular}}

\ifCLASSOPTIONcompsoc
  \usepackage[nocompress]{cite}
\else
  \usepackage{cite}
\fi

\usepackage[numbers]{natbib}

\ifCLASSINFOpdf
\else
\fi
\begin{document}

\sloppy

%
\title{FedEval: A Holistic Evaluation Framework for Federated Learning}
%
%
%
%

\author{Di~Chai$^\dag$, Leye~Wang$^\dag$, Liu~Yang, Junxue~Zhang, Kai~Chen, Qiang~Yang,~\IEEEmembership{Fellow,~IEEE}
\thanks{$^\dag$Equal contribution, ranked alphabetically}
\IEEEcompsocitemizethanks{
\IEEEcompsocthanksitem Di~Chai, Liu~Yang, Junxue~Zhang, Kai~Chen, and Qiang~Yang are with Hong Kong University and Science and Technology, Hong Kong SAR, China\protect\\
E-mail: dchai@cse.ust.hk, lyangau@cse.ust.hk, jzhangcs@cse.ust.hk, kaichen@cse.ust.hk, qyang@cse.ust.hk 
\IEEEcompsocthanksitem Leye Wang is with Key Lab of High Confidence Software Technologies, Peking University, China.\protect\\
E-mail: leyewang@pku.edu.cn
}
\thanks{Manuscript received Jun 30, 2022; revised Jun 30, 2022.}}

%
%

\markboth{Journal of \LaTeX\ Class Files,~Vol.~14, No.~8, August~2015}%
{Shell \MakeLowercase{\textit{et al.}}: Bare Demo of IEEEtran.cls for Computer Society Journals}

%



\IEEEtitleabstractindextext{%
\begin{abstract}

Federated Learning (FL) has been widely accepted as the solution for privacy-preserving machine learning without collecting raw data. While new technologies proposed in the past few years do evolve the FL area, unfortunately, the evaluation results presented in these works fall short in integrity and are hardly comparable because of the inconsistent evaluation metrics and experimental settings. In this paper, we propose a holistic evaluation framework for FL called FedEval, and present a benchmarking study on seven state-of-the-art FL algorithms. Specifically, we first introduce the core evaluation taxonomy model, called FedEval-Core, which covers four essential evaluation aspects for FL: Privacy, Robustness, Effectiveness, and Efficiency, with various well-defined metrics and experimental settings. Based on the FedEval-Core, we further develop an FL evaluation platform with standardized evaluation settings and easy-to-use interfaces. We then provide an in-depth benchmarking study between the seven well-known FL algorithms, including FedSGD~\cite{mcmahan2016communication}, FedAvg~\cite{mcmahan2016communication}, FedProx~\cite{FedProx}, FedOpt~\cite{FedOpt}, FedSTC~\cite{FedSTC}, SecAgg~\cite{SecureAggregation}, and HEAgg~\cite{aono2017privacy,chai2020secure}. We comprehensively analyze the advantages and disadvantages of these algorithms and further identify the suitable practical scenarios for different algorithms, which is rarely done by prior work. Lastly, we excavate a set of take-away insights and future research directions, which are very helpful for researchers in the FL area.

\end{abstract}

}

\maketitle

\IEEEdisplaynontitleabstractindextext

%
\IEEEpeerreviewmaketitle

\section{Introduction}\label{sec:introduction}

Data privacy is becoming an increasingly severe issue today since more and more real-life applications are data-driven. Companies that fail to protect users' privacy may face a hefty fine, \eg FTC fines Facebook \$5 billion to force new privacy references. \citet{mcmahan2016communication} in Google proposed the idea of federated learning (FL) to meet the privacy-preserving regularization. Instead of collecting a massive amount of user data for machine learning (ML) training, FL sets up a joint training scenario in which the client devices, united by a common agreement under a central authority, participate in the model training and only upload specific model parameters to the cloud server for aggregation.

Since FL was proposed, various research work has appeared, targeting different problems in FL. Briefly, these work could be categorized into heterogeneity-oriented, efficiency-oriented, and security-privacy-oriented solutions. For example, some work used gradient compression technology \cite{pmlr-v119-li20g,FedSTC} and guided client selection \cite{DBLP:conf/iclr/YangFL21,Oort} to improve the efficiency, some work explored using cryptography-based techniques to protect the private data \cite{SecureAggregation,POSEIDON,aono2017privacy,chai2020secure}, and many works focused on solving the heterogeneity issue \cite{DBLP:conf/icml/ShamsianNFC21,DBLP:conf/icml/AcarZZNMWS21,DBLP:conf/icml/CollinsHMS21,DBLP:conf/nips/HanzelyHHR20,FedProx,DBLP:conf/iclr/Diao0T21,DBLP:conf/icml/Dennis0S21}.

While these works focus on one or two issues in FL, their evaluation results are also restricted to the corresponding areas. For example, FedAvg \cite{mcmahan2016communication} tries to reduce the communication rounds by adding the number of clients' local updates, but the resulting increased local running time is not evaluated. Moreover, the non-IID issue is also not thoroughly tested in their work. \citet{zhao2018federated} used a shared dataset to solve the non-IID data problem, but they dismissed the potential private data leakage caused by the shared dataset. Analysis of these existing works raises several questions. \textbf{\textit{Is it adequate to only evaluate the aspects in which the progress is made?} \textit{What are the metrics that need to be considered in the evaluation of FL systems?} \textit{How can we compare existing FL technologies if the reported evaluation results focus on different aspects?}}

To illustrate the need for a well-designed evaluation model, we now show two concrete examples of the ambiguity caused by inappropriate evaluation metrics.

\textbf{Is a smaller number of communication rounds always better?} Most of the existing works, including Google's FedAvg \cite{mcmahan2016communication}, only use the number of communication rounds to measure efficiency. However, the time consumption per round grows because of the rising workload in local updates. In other words, the time of local updates is also essential in computational efficiency but neglected in some prior work \cite{mcmahan2016communication,FedSTC,FedOpt,FedProx}. Thus we can hardly say that a smaller number of communication rounds is identical to less time consumption, especially when the model is large and requires more hardware resources.

\textbf{How to properly measure the private data leakage from intermediate results?} Most FL studies follow the workflow of the FedAvg \cite{mcmahan2016communication} algorithm, which assumes that it is secure and private to exchange the model parameters in plaintext while keeping the data locally. However, recent works have shown that attackers can recover the data from gradients in some cases \cite{aono2017privacy, zhu2019deep}. Thus it is essential to properly assess and compare the private data leakage of different algorithms. Existing work has tried to measure or analyze privacy preservation using differential privacy (DP), homomorphic encryption (HE), random masks, \etc. However, these analyses and measurements are not comparable because different definitions and assumptions are used. Thus, it is essential and urgently required to standardize the evaluation of private data leakage in FL and produce comparable results.

To answer these questions, standardize the assessment, and encourage healthy developments of FL, we propose a holistic evaluation framework called \fedeval. \fedeval has a well-designed core evaluation taxonomy model, called \fedeval-Core, that contains four important evaluation targets in FL: \textbf{privacy}, \textbf{robustness}, \textbf{effectiveness}, and \textbf{efficiency}. For each target, we propose detailed evaluation methodologies in \S\ref{sec:method}. These evaluation targets are usually in the form of tradeoffs to each other, \eg enhancement of privacy preservation usually brings the loss of efficiency~\cite{aono2017privacy}, thus a comprehensive evaluation is essential to fairly assess FL models. Moreover, adopting the holistic evaluation gives us insight into finding new research topics by combining different evaluation targets, which is discussed in \S\ref{sec:insights}.

Another hindrance to the standardized and comprehensive evaluation of FL systems is the lack of a common evaluation platform. Users usually need to first implement the evaluation metrics and gather the results manually. Different evaluation settings (\eg the way of collecting the communication statistics\footnote{For example, the amount of transmitted data can be estimated by measuring the size of exchanging variables, which can be done using packages. However, different packages may yield different results.}) adopted in different papers result in non-comparable evaluation results. Thus, a common evaluation platform with a built-in evaluation model is also the need of the day. 

To solve this problem, we have designed and implemented a lightweight and easy-to-use evaluation platform in \fedeval using the proposed evaluation taxonomy model. Briefly, our platform provides an evaluation service through which the users could easily evaluate new FL algorithms or test new attack and defense methods.

Based on \fedeval, we have evaluated seven FL algorithms, FedSGD, FedAvg, FedProx, FedSTC, FedOpt, SecAgg, and HEAgg. The evaluation results show that these FL algorithms have both advantages and disadvantages, and we further identify the suitable practical scenarios for different algorithms. We also have formulated the evaluation results into radar charts for easy comparisons. Based on the evaluations, we further excavate a group of insights and future directions regarding improving FL algorithms for the researchers. \fedeval is fully open-sourced\footnote{\url{https://github.com/Di-Chai/FedEval}}, and we will keep updating the framework to provide better evaluation services. We believe that \fedeval can significantly benefit the development of FL by providing standardized and comparable evaluation results, easy-to-use evaluation tools, and guidance for designing new FL algorithms.

We have noted that there exists studies that focuses on FL benchmarks~\cite{caldas2018leaf,hu2020oarf,DBLP:conf/middleware/NilssonSUGJ18,liu2020evaluation} and FL platforms~\cite{FATE,fedscale}. Our work differs from these studies in the following aspects: 1) Compared with FL benchmarks studies, our solution provides an evaluation platform that can significantly standardize the benchmarks and produce more comprehensive evaluations, which are discussed in \S\ref{sec:related_work}. 2) Compared with existing FL platforms (\eg, FedScale~\cite{fedscale} and FATE~\cite{FATE}), which devote their effort to improving the usability of implementing FL algorithms, although the evaluation metrics are largely covered by these platforms or could be added in the future, the evaluation metrics are not well organized and the evaluation processes are not standardized. Our study focuses on properly organizing the evaluation metrics and standardizing the evaluation process by proposing the \fedeval-Core evaluation taxonomy model, which enables us to evaluate and analyze FL algorithms from a new perspective. Moreover, \fedeval could also be transferred from evaluating the FL algorithms to evaluating the FL platforms\footnote{Even for the same FL algorithm, different platforms may have diverse implementations, leading to different efficiency, effectiveness, robustness, and privacy preservation.} in the future, which is further discussed in \S\ref{sec:extensibility}.

\vspace{-2mm}
\section{\fedeval-Core}\label{sec:method}
\vspace{-2mm}

\begin{figure}[!h]
	\centering
	\includegraphics[width=0.5\textwidth]{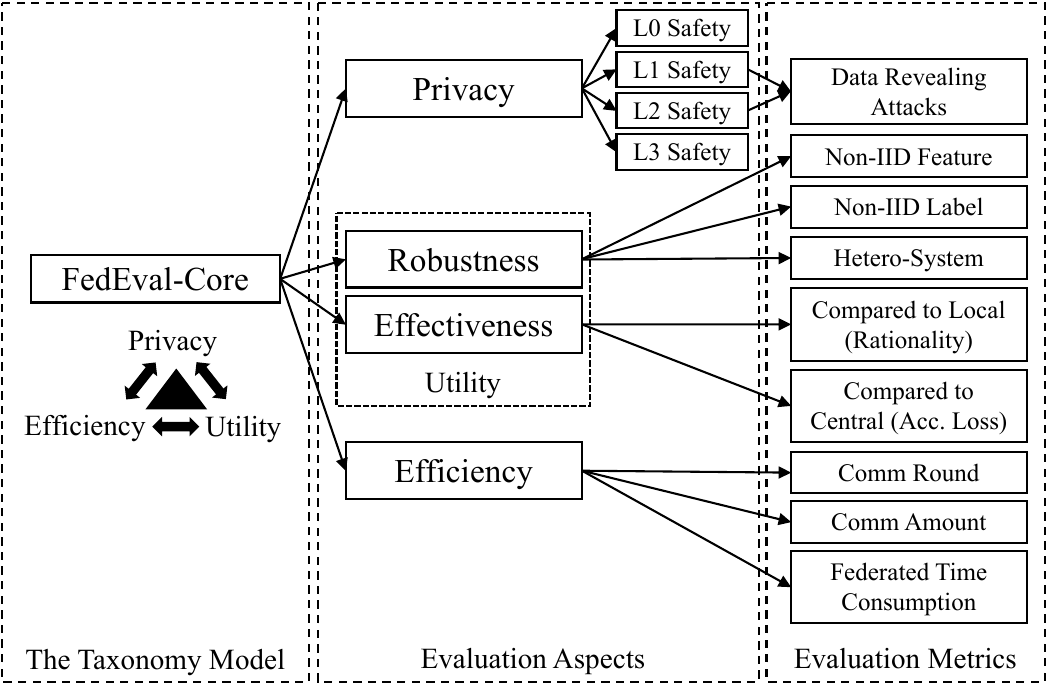}
	\vspace{-5mm}
	\caption{The evaluation taxonomy model.}
	\label{fig:taxonomy}
\end{figure}
	
In this section, we present the evaluation taxonomy model called \fedeval-Core, which is illustrated in \Cref{fig:taxonomy}. We design the \fedeval-Core taxonomy based on a tradeoff triangle of privacy, utility, and efficiency in FL \cite{NFL}. For clarity, we further split the utility into two sub-objectives: robustness which describes the utility loss caused by the external heterogeneity, and effectiveness which describes the utility loss caused by the internal design (\eg, adopting differential privacy). This tradeoff triangle could be used to explain most of the FL studies. For example, FedAvg improves the efficiency but suffers from utility loss; studies adopt more complex optimization targets to solve the heterogeneity problem resulting in lower efficiency; encryption-based FL solutions improve privacy preservation but have low efficiency, differential-privacy-based FL solutions have the tradeoff between privacy and utility, \etc.

In the rest of this section, we introduce each part of the \fedeval-Code taxonomy model in detail.

\subsection{Privacy Evaluation}

Privacy preservation is the foundation of FL. FL is proposed to enable joint machine learning model training without revealing private data. The most common FL architecture used by FL studies is that all the clients (\ie, data owner) perform local training and only upload intermediate results (\eg, model gradients), which will be subsequently aggregated by a central server. The aggregated results will be distributed to the clients for the next round of training. During the FL training process, both the server and clients could be adversaries who want to reveal private data.

The most rigorous way to show privacy preservation of FL algorithms is by giving security proof. However, according to our investigation of top venues\footnote{We investigated 60+ FL papers published from Jun 2017 to Jun 2021 on NeurIPS, ICML, ICLR, KDD, CCS, NDSS, OSDI, \etc.}, less than 10\% of FL studies provided rigorous proof showing their methods are with semantic security or satisfy differential privacy (DP). Intuitively, the ratio is low because 1) It is hard to provide security proof for an FL protocol/algorithm unless DP or secure multi-party computation (SMPC) technologies are adopted to protect the entire FL process; 2) Utilizing the DP or SMPC (\eg, homomorphic encryption) may bring high model utility loss or efficiency overhead, respectively.

We propose to evaluate the privacy preservation of the FL algorithm/system through empirical studies. We think it is an excellent supplementary evaluation method for rigorous proof, especially for FL studies which are hard to analyze the security according to cryptography definitions.

First, we propose a categorization that classifies FL studies into different security levels according to how much information the adversary learns, which is presented in \Cref{tab:security_level}. The conventional centralized data collection method has level-0 safety because all the data are directly exposed, and all the FL methods satisfy level-0 safety. Most FL work is level-1 safety because the server knows both individual and aggregated gradients. Afterward, secure aggregation \cite{SecureAggregation} is proposed to hide the individual intermediate results, and the server can learn the aggregated parameters, which is classified as level-2 safety. Level-3 safety is achieved when the server learns nothing during the FL training. FL work adopting fully homomorphic encryption, \eg, Poseidon~\cite{POSEIDON}, could achieve level-3 safety.

Second, we quantitively compare the FL framework in level-1 and level-2 security by measuring their ability to defend the existing data-revealing attacks. The quantitive analysis for the FL framework in level-0 and level-3 security is unnecessary because they either leak all the private data or leak nothing. Regarding the data-revealing attacks, we currently adopt the gradients attacks. Recent work \cite{aono2017privacy} has proved that the gradients are proportional to the input data in fully connected (FC) layers. Using linear regression as an example, the loss function and parameters updating function are $l = ||\mathbf{wx}+b-y||^2_2, \partial l / \partial \mathbf{w} = 2 \mathbf{x}^T(\mathbf{wx}+b-y)$, where $\mathbf{x} \in \mathbb{R}^{m}$, $\mathbf{w} \in \mathbb{R}^{m}$, $m$ is the data dimension, $b$ is scalar representing the bias, and $y$ is also scalar representing the regression target. Since $(\mathbf{wx}+b-y)$ is scalar, $\partial l / \partial \mathbf{w}$ is proportional to input data $\mathbf{x}$. Thus, the adversary can easily recover $\mathbf{x}$ by rescaling $\partial l / \partial \mathbf{w}$, which is denoted as FCAttack in our paper. Gradients attacks on deep models, called deep leakage from gradients (DLG)~\cite{zhu2019deep}, are also proposed. Briefly, the attackers replace input data and labels with random initialized variables, then train them using the gradients. The main idea is that if the gradients produced by the fake data and fake labels are close enough to the true gradients under the same model weights, then the fake data and fake labels will be close to the true data and true labels. For now, we have implemented these two types of attacks in our evaluation system, and we will keep adding more powerful attack methods with the development of FL.

\begin{table}[h!]
	\centering
	\renewcommand\arraystretch{0.6}
	\setlength{\tabcolsep}{0.3em}
	\vspace{2mm}
	\caption{A classification of different security levels.}
	\label{tab:security_level}
	\vspace{-2mm}
	\begin{tabular}{c|c|c|c}
		\toprule
		\tabincell{c}{Safety\\Levels} & Definition & Representative Work & Attack \\
		\midrule
		Level 3 & \tabincell{c}{Adversary only knows\\encrypted messages} & Poseidon~\cite{POSEIDON} & $\backslash$ \\
		\midrule
		Level 2 & \tabincell{c}{Adversary knows\\aggregated messages\\in plaintext} & \tabincell{c}{Secure\\Aggregation~\cite{SecureAggregation}} & \multirow{4}{*}{\cite{aono2017privacy},\cite{zhu2019deep}, \etc.} \\
		\cmidrule(l){1-3}
		Level 1 & \tabincell{c}{Adversary knows\\individual messages\\in plaintext} & \tabincell{c}{Most FL studies. \\ FedAvg, FedProx, \etc.} & \\
		\midrule
		Level 0 & \tabincell{c}{Gather data centrally} & \tabincell{c}{All FL work \\ achieve higher level} & NA \\
		\bottomrule
	\end{tabular}
\end{table}

\vspace{-4mm}
\subsection{Robustness Evaluation}

The FL training usually happens across distributed devices, which poses many uncertainties in the systems. The uncertainties are usually categorized into the following two types:

\parab{Statistical uncertainties:} FL aims at fitting a model to data generated by different participants. Each participant collects the data in a non-IID manner across the network. The amount of data held by each participant may also significantly differ. The non-IID data issue poses challenges to the training of FL. The model will be more difficult to train under such abnormal data distribution. The non-IID data problem could be further categorized \cite{ma2022state} into: 

\begin{icompact}
	\item Non-IID feature setting: The $P(y|x)$ of different parties are the same while the $P(x)$ are different. For example, in the FEMNIST dataset, different clients hold the same label space containing the same set of symbols but they have different hand-writing styles on the same symbols.
	\item Non-IID label setting: The $P(x|y)$ of different parties are the same while the $P(y)$ are different. For instance, in the MNIST dataset, the non-IID data is usually simulated by allocating different labels to different parties~\cite{mcmahan2016communication} such that $P(y)$ are different while the feature distributions under the same label are the same.
\end{icompact}

Both non-IID feature and label settings are used in existing work for robustness evaluation. However, existing work usually does not differentiate these two settings and different settings might be adopted in the evaluation, which is not reasonable because these two settings may have different impacts on the model performance. Thus, in our evaluation framework, we clearly specify the non-IID setting and report both evaluation results.

\parab{System uncertainties:} A large number of participants indicate a significant disparity on devices. Different networks (3G, 4G, WiFi), storage, computing resources (CPU, GPU), and power (using battery, charging) may exist in the system. These system uncertainties could cause issues like stragglers and dropouts (\eg, out of battery) \cite{smith2017federated}.

We define the robustness evaluation as the testing under non-IID data, system stragglers, and dropouts. A robust FL system should be able to perform consistently well given such uncertainties.

\vspace{-3mm}
\subsection{Effectiveness Evaluation}

In the effectiveness metric, we care about the performance of the obtained machine learning model (\ie, the predictive power). Adequate data is usually an indispensable condition to achieving satisfactory ML accuracy, especially when deep learning is applied. However, such a condition usually cannot be satisfied in the real world due to privacy-preserving restrictions. Each data owner can only access its local data, which is also known as the isolated data islands problem \cite{yang2019federated}. FL systems should be able to break such isolation and achieve performance (FLEffectiveness) better than the LocalEffectiveness, \ie, training model locally without joining any federations.

In federated learning, we typically learn the global model by solving the following problem:

\vspace{-2.5mm}
\begin{equation} \label{eq:fl_target}
	\min_w f(w) = \sum_{k=1}^N p_k F_k(w) = \mathbb{E}_k[F_k(w)]
\end{equation}
\vspace{-2.5mm}

where $N$ is the number of clients, $p_k \ge 0$ and $\sum_k p_k = 1$. $F_k(w)$ is defined as the empirical loss over the local data samples, \ie, $F_k(w)=\frac{1}{n_k}\sum_{i=1}^{n_k}l_i(w)$ \cite{li2019fair}, where $n_k$ is the number of samples at the $k$-th party, and we set $p_k=\frac{n_k}{n}$ where $n=\sum_k n_k$ is the total number of samples.

\begin{definition}[FLEffectiveness~(FE)] \label{def:flacc}
	We define the FL effectiveness as : $ \sum_{k=1}^N p_k Acc(h(w,x_k),y_k)$, where $w$ is the model parameter learned from Equation \ref{eq:fl_target}, $h(w,x_k)$ outputs a probability distribution over the classes or categories that can be assigned to $x_k \sim D_k$, $Acc$ function computes accuracy of $h(w,x_k)$ regarding the label $y_k$, and we set $p_k=\frac{n_k}{n}$.
\end{definition}

\begin{definition}[LocalEffectiveness~(LE)]
	Using the same notation in Definition \ref{def:flacc} , we define the LocalEffectiveness as: $ \sum_{k=1}^N p_k Acc(h(w_k,x_k),y_k)$, where $w_k$ is the local model parameter learnt by minimizing the local objective: $w_k=\arg\min_{w} F_k(w)$, and we set $p_k=\frac{n_k}{n}$.
\end{definition}

\begin{definition}[CentralEffectiveness~(CE)]\footnote{Centralized data collection and training is only an ideal experimental situation that represents a theoretical accuracy upper bound. In reality, we usually cannot put all the data in one place due to the restriction of privacy regulations.}
We define the CentralEffectiveness as $Acc(h(w,x), y)$, where $w$ is the model parameter trained by $min_w F(w) :=\mathbb{E}_{x \sim D}[f(w,x)]$, $x$ represents data that collected from all the clients, and $D$ is the global data distribution.
\end{definition}

We compare the FE and LE to measure how much the model performance is improved using federated learning.

Meanwhile, FL systems should be able to obtain approximately the same accuracy as that of centralized machine learning systems. In other words, FE should be bounded: FE$\le$CE, which can be easily proved because the centralized training can simulate any form of federated training and the central training is also exempt from the performance drop caused by the statistical and system uncertainties, which is the main reason of the performance reduction of FL.

Assuming that CE is significantly higher than LE, we can have the following analysis: If LE$\ge$FE, then the FL system has failed to converge. If FE$\approx$CE, then the FL system shows no accuracy decline, which is the best case. Given two FL systems, $FL_1$ and $FL_2$, if $FE_1 > FE_2$, then we say $FL_1$ has better effectiveness than $FL_2$.

\vspace{-3mm}
\subsection{Efficiency Evaluation}

Efficiency evaluation is essential for the FL mechanism because communication and computation are usually the bottlenecks in the applications. Furthermore, FL typically happens across different devices or different data centers, which have relatively worse networking conditions (\eg, small bandwidth and high delay) than centralized model training. Meanwhile, FL adopts privacy-preserving techniques (\eg, HE), which bring significant computation or communication burden to the system. Thus an efficiency evaluation is indispensable in the FL assessments.

\textbf{Comunication} and \textbf{Time} are two frequently used metrics in the efficiency evaluation of FL. We measure the communication in two aspects: the number of communication rounds ($CommRound$) and the total amount of data transmission during training ($CommAmount$). $CommRound$ is related to the convergence speed. For the time consumption metric, we measure the total time required to finish the FL training under the same stopping criteria.

\begin{figure}[!h]
	\vspace{-2mm}
	\centering
	\begin{subfigure}[t]{0.24\textwidth}
		\includegraphics[scale=0.55]{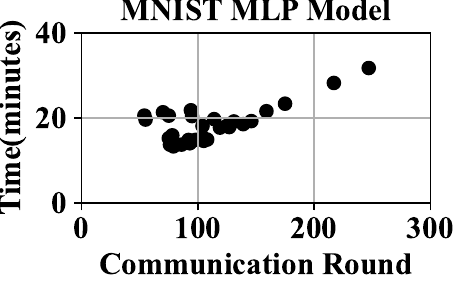}
	\end{subfigure}
	\begin{subfigure}[t]{0.24\textwidth}
		\includegraphics[scale=0.55]{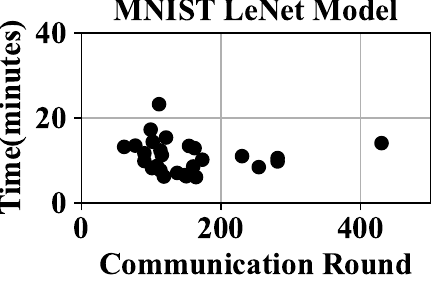}
	\end{subfigure}
	\vspace{-5mm}
	\caption{Small communication rounds are not always identical to less time consumption.}
	\label{fig:time_vs_comm}
\end{figure}

\parab{Why time evaluation cannot be dismissed in FL.} Most of the FL studies only use communication rounds as the metric to measure efficiency. The time consumption is usually overlooked in the evaluation. We have collected the communication rounds and time consumption in multiple times of experiments under different parameters, and the results are presented in \Cref{fig:time_vs_comm}. We can conclude that the communication rounds are not always linearly correlated with the time consumption. When the model has the minimum communication rounds, it tends to have large time consumption. Thus the time consumption cannot be dismissed in the evaluation of FL.



\label{sec:federated_time}
\parab{Time evaluation in a federated manner.} Federated learning tends to have many participants in the application, \eg, Google's federated mobile keyboard prediction~\cite{mcmahan2016communication} may have million-scale participants. Thus it is necessary to evaluate the efficiency in such a large-scale participants scenario. The best solution for evaluating large-scale participants is maintaining and running experiments directly on a huge number of devices, and each device represents a client. However, few institutions have the ability to maintain and perform evaluations on such a huge amount of devices, and it is extremely not practical for individual researchers. Thus, the practical solution which is adopted by most FL studies is simulating all the clients using a few computing servers. However, simulating a large number of clients using limited computing servers may have biased time evaluation results due to resource competition. Our solution is performing the time evaluation in a federated manner, limiting the maximum computing resources occupied by each client and the maximum number of clients the system can simultaneously support. For the time consumption metric, we collect the federated time consumption rather than the real-world time consumption. For instance, if the system can simultaneously support ten clients (\eg, the CPU has ten cores) and the algorithm selects 100 clients in each round of training, our framework executes the clients ten by ten, collects the individual time consumption of each client, and outputs the maximum individual time as the federated time consumption in this round of training.

\begin{table*}[!h]
	\centering
	\renewcommand\arraystretch{1}
	\setlength{\tabcolsep}{1.0em}
	\scriptsize
	\vspace{+2mm}
	\caption{An investigation of evaluation metrics used in representative FL work. The "\CIRCLE" symbol is well evaluated, and "\LEFTcircle" is partially evaluated. The empty entries represent that the metrics are not evaluated.}
	\vspace{-2mm}
	\label{tab:used_metrics}
	\begin{tabular}{l|c|c|c|c|c|c|c|c}
		\hline
		Literature & \tabincell{c}{Privacy} & \tabincell{c}{Non-IID\\Feature} & \tabincell{c}{Non-IID\\Label} & \tabincell{c}{Hetero\\System} & \tabincell{c}{Comm\\Round} & \tabincell{c}{Comm\\Amount} & \tabincell{c}{Time} & \tabincell{c}{Effectiveness} \\
		\hline
		\citet{DBLP:conf/nips/SmithCST17} (\textit{NeurIPS 17}) & & \CIRCLE &  & \CIRCLE &  &  & \CIRCLE & \CIRCLE \\
		\hline
		\citet{DBLP:conf/ccs/BonawitzIKMMPRS17} (\textit{CCS 17}) & \CIRCLE &  &  & \CIRCLE &  & \CIRCLE & \CIRCLE &  \\
		\hline
		
		\citet{mohri2019agnostic} (\textit{ICML 19}) & &  &  &  &  &  &  & \LEFTcircle \\
		\hline
		\citet{DBLP:conf/icml/YurochkinAGGHK19} (\textit{ICML 19}) & &  & \CIRCLE &  & \CIRCLE &  &  &  \\
		\hline
		
		\citet{DBLP:conf/iclr/PengHZS20} (\textit{ICLR 20}) & & \CIRCLE &  &  &  &  &  & \LEFTcircle \\\hline
		\citet{wang2020federated} (\textit{ICLR 20}) & &  & \CIRCLE &  & \CIRCLE & \CIRCLE &  & \LEFTcircle \\\hline
		\citet{li2019fair} (\textit{ICLR 20}) & & \CIRCLE &  &  & \CIRCLE &  &  & \LEFTcircle \\\hline
		\citet{pmlr-v119-li20g} (\textit{ICML 20}) & &  &  &  &  & \CIRCLE &  & \LEFTcircle \\\hline
		\citet{DBLP:conf/icml/HamerMS20} (\textit{ICML 20}) & &  &  &  & \CIRCLE &  &  & \LEFTcircle \\\hline
		\citet{DBLP:conf/icml/YuRMK20} (\textit{ICML 20}) & &  &  &  &  &  &  & \CIRCLE \\\hline
		\citet{DBLP:conf/icml/RothchildPUISB020} (\textit{ICML 20}) & &  &  &  &  &  &  & \LEFTcircle \\\hline
		\citet{DBLP:conf/icml/KarimireddyKMRS20} (\textit{ICML 20}) & & \CIRCLE &  &  & \CIRCLE &  &  & \LEFTcircle \\\hline
		\citet{DBLP:conf/icml/MalinovskiyKGCR20} (\textit{ICML 20}) & &  &  &  & \CIRCLE &  & \CIRCLE & \LEFTcircle \\\hline
		\citet{DBLP:conf/nips/0001MO20} (\textit{NeurIPS 20}) & &  & \CIRCLE &  &  &  &  & \LEFTcircle \\\hline
		\citet{DBLP:conf/nips/DinhTN20} (\textit{NeurIPS 20}) & &  & \CIRCLE &  & \CIRCLE &  &  & \LEFTcircle \\\hline
		\citet{DBLP:conf/nips/PathakW20} (\textit{NeurIPS 20}) & &  &  &  & \CIRCLE &  &  & \LEFTcircle \\\hline
		\citet{DBLP:conf/nips/MarfoqXNV20} (\textit{NeurIPS 20}) & &  &  & \CIRCLE & \CIRCLE &  & \CIRCLE &  \\\hline
		\citet{DBLP:conf/nips/WangLLJP20} (\textit{NeurIPS 20}) & &  & \CIRCLE &  & \CIRCLE &  &  & \LEFTcircle \\\hline
		\citet{DBLP:conf/nips/ReisizadehFPJ20} (\textit{NeurIPS 20}) & &  & \CIRCLE &  &  &  &  & \LEFTcircle \\\hline
		\citet{DBLP:conf/nips/HanzelyHHR20} (\textit{NeurIPS 20}) & &  &  &  & \CIRCLE &  &  & \LEFTcircle \\\hline
		\citet{DBLP:conf/nips/0001AA20} (\textit{NeurIPS 20}) & &  & \CIRCLE &  & \CIRCLE & \CIRCLE & \CIRCLE & \LEFTcircle \\\hline
		\citet{DBLP:conf/nips/GrammenosMCM20} (\textit{NeurIPS 20}) & \CIRCLE &  &  &  &  &  & \CIRCLE & \LEFTcircle \\\hline
		\citet{DBLP:conf/nips/DaiLJ20} (\textit{NeurIPS 20}) & & \CIRCLE &  &  &  &  & \CIRCLE &  \\\hline
		\citet{DBLP:conf/nips/YuanM20} (\textit{NeurIPS 20}) & &  &  &  & \LEFTcircle &  &  &  \\\hline
		\citet{DBLP:conf/nips/GhoshCYR20} (\textit{NeurIPS 20}) & & \CIRCLE &  &  &  &  &  & \CIRCLE \\\hline
		\citet{DBLP:conf/nips/DubeyP20} (\textit{NeurIPS 20}) & \CIRCLE &  &  &  & \LEFTcircle &  &  & \LEFTcircle \\\hline
		\citet{DBLP:conf/nips/DengKM20} (\textit{NeurIPS 20}) & &  & \CIRCLE &  & \CIRCLE &  & \CIRCLE & \LEFTcircle \\\hline
		\citet{DBLP:conf/nips/LinKSJ20} (\textit{NeurIPS 20}) & &  & \CIRCLE &  & \CIRCLE &  &  & \LEFTcircle \\\hline
		\citet{DBLP:conf/mobicom/NiuWTHJLWC20} (\textit{NeurIPS 20}) & \CIRCLE & \CIRCLE &  & \CIRCLE & \CIRCLE &  &  & \LEFTcircle \\\hline
		\citet{DBLP:conf/kdd/GuDLH20} (\textit{KDD 20}) & \CIRCLE &  &  &  &  &  & \CIRCLE & \LEFTcircle \\\hline
		
		\citet{DBLP:conf/iclr/ZhangSFYA21} (\textit{ICLR 21}) & & \CIRCLE & \CIRCLE &  & \CIRCLE &  &  & \LEFTcircle \\\hline
		\citet{DBLP:conf/iclr/YangFL21} (\textit{ICLR 21}) & &  & \CIRCLE &  & \CIRCLE & \CIRCLE & \CIRCLE & \LEFTcircle \\\hline
		\citet{FedOpt} (\textit{ICLR 21}) & &  &  &  & \CIRCLE &  &  & \LEFTcircle \\\hline
		\citet{DBLP:conf/iclr/ChenC21} (\textit{ICLR 21}) & &  & \CIRCLE &  &  &  &  & \LEFTcircle \\\hline
		\citet{DBLP:conf/iclr/LiJZKD21} (\textit{ICLR 21}) & & \CIRCLE &  &  & \CIRCLE &  &  & \LEFTcircle \\\hline
		\citet{DBLP:conf/iclr/AcarZNMWS21} (\textit{ICLR 21}) & &  & \CIRCLE &  &  & \CIRCLE &  &  \\\hline
		\citet{DBLP:conf/iclr/Al-ShedivatGXR21} (\textit{ICLR 21}) & &  &  &  & \CIRCLE &  &  & \LEFTcircle \\\hline
		\citet{DBLP:conf/iclr/JeongYYH21} (\textit{ICLR 21}) & &  & \CIRCLE &  & \CIRCLE &  &  & \LEFTcircle \\\hline
		\citet{DBLP:conf/iclr/YoonSHY21} (\textit{ICLR 21}) & \LEFTcircle & \CIRCLE & \CIRCLE &  & \CIRCLE &  &  & \CIRCLE \\\hline
		\citet{DBLP:conf/iclr/Diao0T21} (\textit{ICLR 21}) & & \CIRCLE & \CIRCLE &  &  &  &  & \LEFTcircle \\\hline
		\citet{DBLP:conf/icml/MurataS21} (\textit{ICML 21}) & &  & \CIRCLE &  &  &  &  & \LEFTcircle \\\hline
		\citet{DBLP:conf/icml/FraboniVKL21} (\textit{ICML 21}) & &  &  &  & \CIRCLE &  &  & \LEFTcircle \\\hline
		\citet{DBLP:conf/icml/AcarZZNMWS21} (\textit{ICML 21}) & &  & \CIRCLE &  & \CIRCLE &  &  &  \\\hline
		\citet{DBLP:conf/icml/KairouzL021} (\textit{ICML 21}) & \CIRCLE & \CIRCLE & \CIRCLE &  & \CIRCLE &  &  & \LEFTcircle \\\hline
		\citet{DBLP:conf/icml/ShamsianNFC21} (\textit{ICML 21}) & & \CIRCLE & \CIRCLE &  &  &  &  & \LEFTcircle \\\hline
		\citet{DBLP:conf/icml/Dennis0S21} (\textit{ICML 21}) & &  & \CIRCLE &  & \CIRCLE &  &  & \LEFTcircle \\\hline
		\citet{DBLP:conf/icml/HuangL0021} (\textit{ICML 21}) & &  & \CIRCLE &  & \CIRCLE &  &  &  \\\hline
		\citet{DBLP:conf/icml/YuanGXYY21} (\textit{ICML 21}) & &  & \CIRCLE &  & \CIRCLE &  &  & \LEFTcircle \\\hline
		\citet{DBLP:conf/icml/CollinsHMS21} (\textit{ICML 21}) & &  & \CIRCLE &  & \CIRCLE &  &  & \LEFTcircle \\\hline
		\citet{DBLP:conf/osdi/LaiZMC21} (\textit{OSDI 21}) & & \CIRCLE &  & \CIRCLE & \CIRCLE &  & \CIRCLE & \LEFTcircle \\\hline
		\citet{DBLP:conf/ndss/SavPTFBSH21} (\textit{NDSS 21}) & \CIRCLE &  & \CIRCLE &  & \CIRCLE &  & \CIRCLE & \CIRCLE \\
		\hline
	\end{tabular}
	\vspace{-4mm}
\end{table*}

\vspace{-2mm}
\subsection{Evaluation Metrics Used in Existing Works}

We have investigated papers on top venues and collected the evaluation metrics used in the representative papers, and \Cref{tab:used_metrics} shows the results. We can observe that 1) All the evaluation aspects are covered in existing work, but they are rarely covered together. However, comprehensive evaluations or analyses are necessary since most of the FL algorithms which make progress in one or two aspects have unpredictable impacts on the performance of the other aspects. For instance, enhancement of privacy preservation may decrease efficiency and effectiveness, improved efficiency may downgrade privacy protection, \etc. Thus we highly recommend FL studies that target proposing new FL algorithms to perform a comprehensive evaluation to unearth the algorithms' advantages and disadvantages thoroughly. Alternatively, the analysis should be provided showing why the evaluation aspects could be omitted; 2) Different non-IID settings are adopted in the robustness evaluation, \ie, non-IID features and non-IID labels, making the evaluation results inconsistent and not comparable; 3) Most studies have non-comprehensive effectiveness evaluations, and comparison to centralized training is usually missing. However, comparison with the centralized model cannot be dismissed since it helps to measure the effectiveness loss of the FL algorithm.


\vspace{-2mm}
\section{The Evaluation Platform} \label{sec:benchmarking_system}

\begin{figure*}[!h]
	\centering
	\includegraphics[scale=0.8]{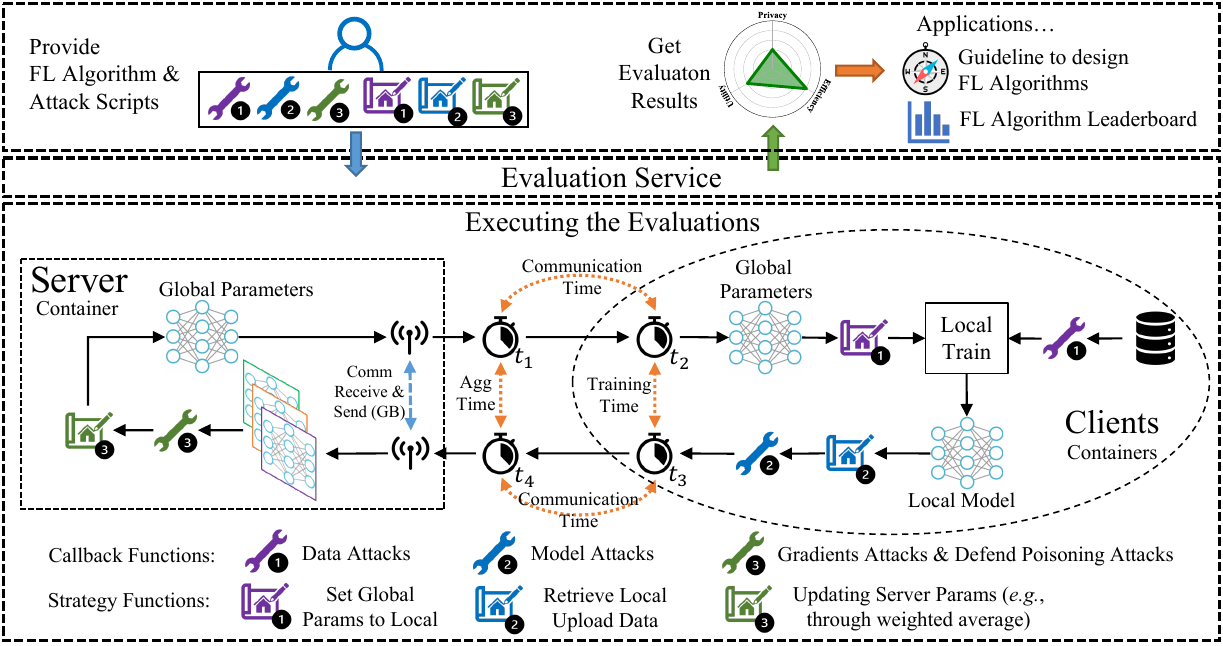}
	\vspace{-3mm}
	\caption{An overview of the FedEval evaluation platform.}
	\label{fig:bm_system_framework}
\end{figure*}

To standardize and simplify the evaluation of FL algorithms, we build an evaluation platform based on \fedeval-Code proposed in \S\ref{sec:method}. \Cref{fig:bm_system_framework} shows an overview of our evaluation platform. Briefly, the users only need to provide a single script containing the necessary federated learning functions or callback functions, \eg, how the server aggregates the parameters from different clients, to evaluate a new FL algorithm or test new attack and defense methods. In the rest of this section, we introduce the architecture and interface of the platform in \S\ref{sec:framework_architecture} and \S\ref{sec:framework_interface}, respectively. 

\vspace{-2mm}
\subsection{Architecture} \label{sec:framework_architecture}

Our evaluation platform has two layers in architecture. The first layer is the evaluation service layer, which provides an interface for users to provide scripts (\eg, a new FL algorithm) and outputs the evaluation results. The evaluation service layer could be used locally or deployed as an online service\footnote{The open-sourced code only supports local usage for now, and the code for deploying online service will be open-sourced in the future.}. The second layer is the execution layer that is responsible for running the assessments based on the \fedeval-Core and returns the results. 

Next, we would like to introduce three challenges of evaluating FL algorithms from the architecture design aspect and our solutions to solve these challenges.

\parab{Participants and Network Simulation.} A widely-used method for simulating multiple participants is using multiprocessing and we think it has the following problems: 1) It is hard to control the hardware resources (\eg, CPU and memory) used by each process; 2) It is hard to evaluate the performance under different network settings (\ie, bandwidth and latency). Our solution is putting all the participants into different docker containers, in which the hardware resources used by each participant could be fully controlled, including the CPU, GPU, memory, disk storage, \etc. The server and clients from different containers communicate through WebSocket. The communication between containers is bridged by container networks. Under such architecture design, it is easy to change the network settings (\ie, bandwidth and latency) by directly configuring the virtual network interface card (NIC).

\parab{Time Evaluation.} The implementation of time evaluation in FL is challenging because it may have many variations based on different purposes. For example, apart from the overall time consumption in each training round, we would also like to provide other time consumption statistics to help the users improve the FL algorithms, \eg, the computation and communication time of the clients, the aggregation time at the server, \etc. The naive implementation of these time evaluation metrics is complicated and requires significant modifications to the platform's source code. Our solution is providing a flexible time evaluation by collecting a group of timestamps, through which multiple time evaluation metrics could be calculated. Specifically, as illustrated in \Cref{fig:bm_system_framework}, we put four timestamps in the platform, which are the time of server send parameters ($t_1$), clients receive parameters ($t_2$), clients send parameters ($t_3$), and server receives parameters ($t_4$). Assuming we have $k$ clients in the training, then $\{ (t_1^i, t_2^i, t_3^i, t_4^i) | 1 \le i \le k \}_n$ represents all the timestamps collected in the $i$-th round. Different combinations of these timestamps have different meanings:

\begin{icompact}
	\item Client computation time (average): $\frac{1}{k}\sum_{i=1}^k (t_3^i - t_2^i)$.
	\item Server aggregation time in the $n$-th round:\\ $sa=min(\{t_1^i | 1 \le i \le k \}_n)-max(\{t_4^i | 1 \le i \le k \}_{n+1})$
	\item Real-world time consumption in the $n$-th round:\\ $min(\{t_1^i | 1 \le i \le k \}_n)-min(\{t_1^i | 1 \le i \le k \}_{n+1})$
	\item Federated time consumption in the $n$-th round:\\ $sa + max(\{t_4^i-t_1^i | 1 \le i \le k \}_{n})$
\end{icompact}

Our platform records all the timestamps and outputs the real-world and federated time consumption. The users are able to compute more metrics based on these timestamps.

\parab{Communication Evaluation.} Communication size is an essential evaluation metric for FL algorithms since the participants in FL tend to have limited network bandwidth, and a large communication size may bring significant efficiency overhead. A naive solution for evaluating the communication size, which is used in many existing FL studies, is directly measuring the size of the transmitted objects in the memory, and many utility packages (\eg, the "getsizeof()" function in python) could be used. However, such evaluation implementation may have two issues: 1) Different packages usually have different outputs; 2) Not all the objects could be accurately assessed using this method. To solve these problems, we measure the communication size by directly collecting data from the virtual NIC, which automatically records how much amount of data is sent out and received. Compared with measuring the transmitted data size in memory, our solution is more accurate and significantly reduces the implementation complexity.


\vspace{-4mm}
\subsection{Interface} \label{sec:framework_interface}

To support the evaluation service illustrated in \Cref{fig:bm_system_framework}, the platform's interface should meet the following requirements: 1) The interface should have high generality and be able to support various FL algorithms based on different techniques; 2) The interface should have high usability and be easily used. In this section, we introduce our platform's interface, which is summarized and tested on many FL algorithms. Specifically, we have implemented eight FL algorithms using our interface, including the fundamental FL methods (\ie, FedSGD and FedAvg), gradients-compression-based methods, federated optimizers, heterogeneity-robust methods, algorithms based on homomorphic encryption, and secure aggregation. Briefly, we have two essential interfaces: the FL algorithm interface (\S\ref{sec:interface_fl_algorithm}) and callback function interface (\S\ref{sec:interface_callback_functions}).

\subsubsection{Interface of FL Algorithms} \label{sec:interface_fl_algorithm}

Our interface of FL algorithm provides the following functions for customizing the host's and clients' behaviors: 

\begin{icompact}
	\item Host: update\_host\_params($client\_parameters$)
	\item Host: retrieve\_host\_download\_info()
	\item Host: host\_select\_train\_clients($ready\_clients$)
	\item Host: host\_select\_eval\_clients($ready\_clients$)
	\item Host: host\_exit\_job()
	\item Client: set\_host\_params\_to\_local($host\_parameters$)
	\item Client: fit\_on\_local\_data()
	\item Client: retrieve\_local\_upload\_info()
	\item Client: local\_evaluate()
	\item Client: client\_exit\_job()
\end{icompact}

The interface functions are named straightforwardly based on their functionalities, \eg, update\_host\_params() takes clients' parameters as inputs and updates the host's parameters (\eg, through weighted average). Due to the space limitation, we put the detailed interface description on our online documentation\footnote{\url{https://fedeval.readthedocs.io/en/latest/tutorial/StrategyAndCallbacks.html}\label{fn:interface}}. Users can override one or multiple of these interface functions to implement new FL algorithms, and \Cref{fig:fl_algorithm_example} shows an example of evaluating naive gradients compression using our interface.

\begin{figure}[h]
\centering
\begin{lstlisting}
from FedEval.strategy import FedAvg

class GradientCompression(FedAvg):
	def retrieve_local_upload_info(self):
		dense_params = self.ml_model.get_weights()
		# Only keep the top 10% weights
		return compress(dense_params, sparsity=0.1)
\end{lstlisting}
\vspace{-2mm}
\caption{An example of evaluating naive gradients compression based on FedAvg using our interface.}
\label{fig:fl_algorithm_example}
\end{figure}

\subsubsection{Interface of Callback Functions}  \label{sec:interface_callback_functions}

Besides the function for implementing new FL algorithms, we also provide the interface of callback functions to make the implementation of attacks and defenses easier. Briefly, we have the following callback functions:

\begin{icompact}
	\item on\_host\_aggregation\_begin($client\_parameters$)
	\item on\_host\_exit()
	\item on\_client\_train\_begin($data$, $model$)
	\item on\_client\_upload\_begin($upload\_parameters$)
	\item on\_setting\_host\_to\_local($host\_parameters$)
	\item on\_client\_exit()
\end{icompact}

Similarly, the callback functions are also named straightforwardly based on the functionalities, and the detailed description for each function could be found in our online document\textsuperscript{\ref{fn:interface}}. Users could override one or multiple of these functions to implement different attack and defense methods, \eg, overriding on\_host\_aggregation\_begin() function to perform gradients attack using the $client\_parameters$ and overriding on\_client\_upload\_begin to protect the individual message before uploading (\eg, through differential privacy). \Cref{fig:callback_example} shows an example of evaluating naive model poisoning attacks through uploading random values.

\begin{figure}[h]
\centering
\begin{lstlisting}
import numpy as np
from FedEval.callback import CallBack

class ModelPoisoning(CallBack):
    def on_client_upload_begin(self, model_params):
        return np.random.random(model_params.shape)
\end{lstlisting}
\vspace{-2mm}
\caption{An example of evaluating naive model poisoning attacks through uploading random values.}
\label{fig:callback_example}
\end{figure}

\vspace{-3mm}
\subsection{Extensibility to Evaluate FL Platforms by \fedeval} \label{sec:extensibility}

With the fast development of FL studies, many FL platforms have appeared to support the application of FL algorithms, \eg, FedScale~\cite{fedscale}, FATE~\cite{FATE}, \etc. However, these platforms typically have different architecture, backends, \etc, making it hard to determine which platform should be used. Regarding the same FL algorithm, these FL platforms may have 1) Different accuracy and efficiency because different backends (\ie, computation and communication backends) are adopted; 2) Different privacy preservation because the implementation of the privacy-preserving protocols might be simplified and not strictly follow the original paper to reduce the implementation complexity. Existing work \cite{fedscale} has already shown that the FL platform's design can impact the algorithms' effectiveness and efficiency.

To solve this problem, we can extend \fedeval from evaluating FL algorithms to assessing FL platforms. Specifically, we can use the same metric from \fedeval-Core, containing the privacy, robustness, effectiveness, and efficiency metrics, to evaluate FL platforms. Furthermore, regarding the evaluation platform, we can integrate different FL platforms as different backends (\ie, the execution layer) and provide a hyper-parameter for users to determine which platforms (\eg, FedScale) will be used in the evaluation. We treat the detailed evaluation metric design and evaluation platform implementation for evaluating the FL platforms as our future work, and we believe such an evaluation will significantly benefit the FL studies.

\vspace{-3mm}
\section{Benchmarking Experimental Results}\label{sec:experiments}

\subsection{Experiment Settings}

Based on \fedeval, we evaluate seven well-known FL algorithms: FedSGD~\cite{mcmahan2016communication}, FedAvg~\cite{mcmahan2016communication}, FedProx~\cite{FedProx}, FedSTC~\cite{FedSTC}, FedOpt~\cite{FedOpt}, Secure Aggregation (SecAgg)~\cite{SecureAggregation}, and aggregation based on additive homomorphic encryption (HEAgg)~\cite{aono2017privacy,chai2020secure}. Five datasets are used in the evaluation: MNIST \cite{lecun1998gradient}, FEMNIST \cite{caldas2018leaf}, CelebA \cite{caldas2018leaf}, Sentiment140 \cite{go2009twitter}, and Shakespeare \cite{LEAFData}. For the hyper-parameters of FL algorithms, we directly follow the original paper, \eg, selecting 10\% of clients in each round of training in FedAvg. For the hyper-parameter during the training, we fix the batch size and local training passes and perform a fine-grained parameter tuning on the learning rates for each algorithm on each dataset. Due to the space limitation, we put the detailed parameter setting and hardware resources on our online document\footnote{\url{https://fedeval.readthedocs.io/en/latest/benchmark/benchmark.html}}.

\vspace{-2mm}
\subsection{Privacy Evaluation} \label{sec:SP benchmark}

\parab{Qualitative Evaluation}: Among these seven FL algorithms, FedSGD, FedAvg, FedProx, FedSTC, and FedOpt belong to level-1 security since they expose individual gradients in plaintext to the adversary. SecAgg and HEAgg belong to level-2 security since the adversary only learns the aggregated model weights in plaintext.

\parab{Quantitative Evaluation}: We perform a quantitative benchmarking study between FedSGD and FedAvg regarding defending against the gradients attacks. Because they represent two different FL training paradigms, the FedSGD aggregates parameters from all clients, and each client only performs very few local updates (\ie one round of local training), while the FedAvg, FedProx, FedSTC, and FedOpt only aggregate parameters from a subset of clients and each client run more rounds of local training. Intuitively, the training paradigm of FedAvg has better performance in concealing the private data because the uploaded gradients are generated by multiple rounds of local training.

\begin{figure}[h!]
	\centering
	\begin{subfigure}[t]{0.48\linewidth}
	\centering
		\includegraphics[width=\textwidth]{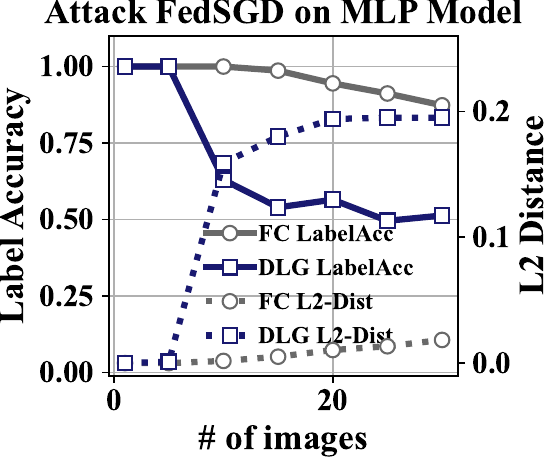}
	\end{subfigure}
	\begin{subfigure}[t]{0.48\linewidth}
	\centering
		\includegraphics[width=\textwidth]{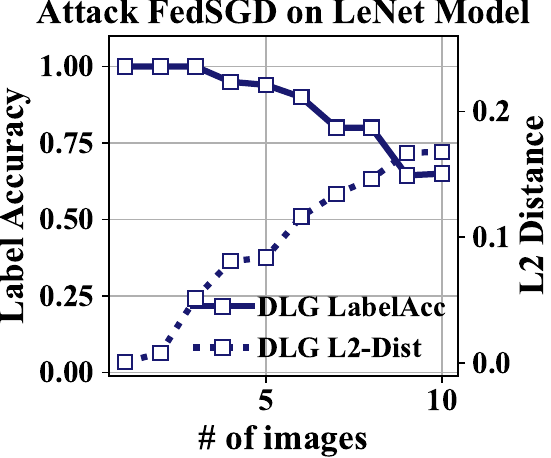}
	\end{subfigure}
	\begin{subfigure}[t]{0.48\linewidth}
	\centering
		\includegraphics[width=\textwidth]{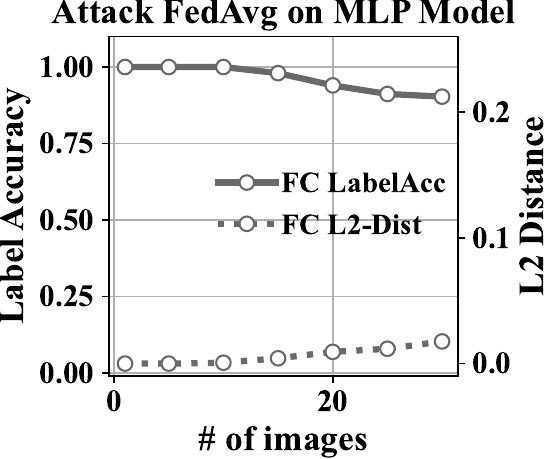}
	\end{subfigure}
	\begin{subfigure}[t]{0.48\linewidth}
	\centering
		\includegraphics[width=\textwidth]{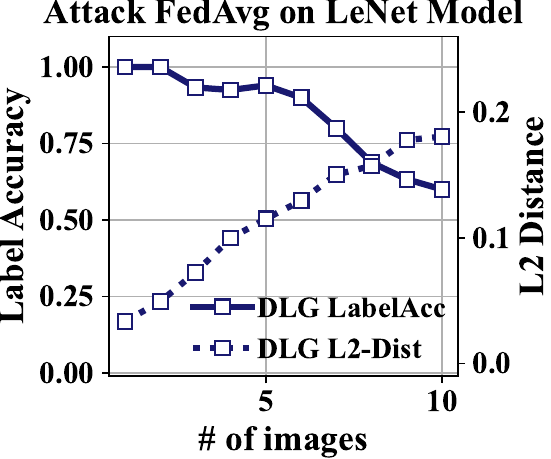}
	\end{subfigure}
	\vspace{-2mm}
	\caption{Vary \# of training images in gradients attack.}
	\label{fig:attack_vary_images}
\end{figure}

\begin{figure}[h!]
	\centering
	\begin{subfigure}[t]{0.48\linewidth}
	\centering
	\includegraphics[scale=0.48]{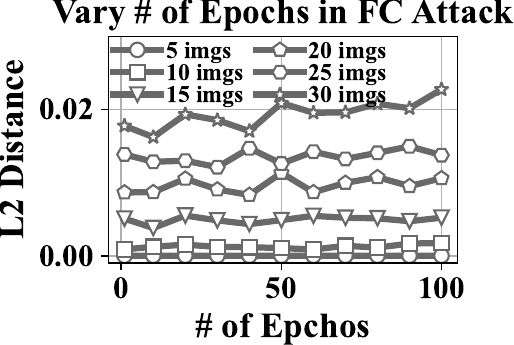}
	\end{subfigure}
	\begin{subfigure}[t]{0.48\linewidth}
	\centering
		\includegraphics[scale=0.48]{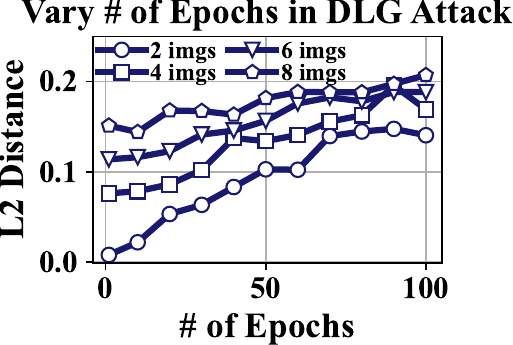}
	\end{subfigure}
	\vspace{-2mm}
	\caption{Vary \# of training epochs in gradients attack.}
	\label{fig:attack_vary_epchos}
\end{figure}

\begin{figure}[h!]
	\centering
	\begin{subfigure}[t]{0.48\linewidth}
		\centering
		\includegraphics[scale=0.48]{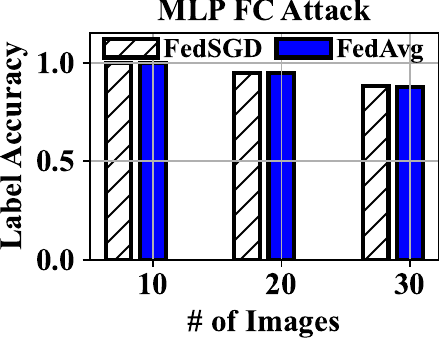}
		\label{fig:compare_fedsgd_fedavg_fc_la}
	\end{subfigure}
	\hspace{-3mm}
	\begin{subfigure}[t]{0.48\linewidth}
	\centering
		\includegraphics[scale=0.48]{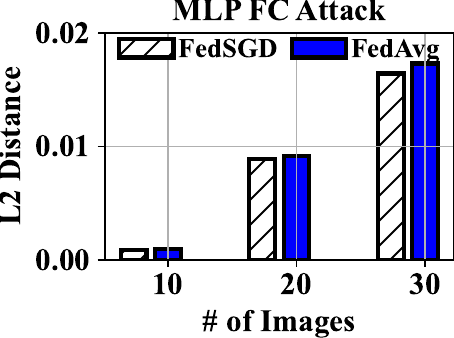}
		\label{fig:compare_fedsgd_fedavg_fc_l2d}
	\end{subfigure}
	\begin{subfigure}[t]{0.48\linewidth}
	\centering
		\includegraphics[scale=0.48]{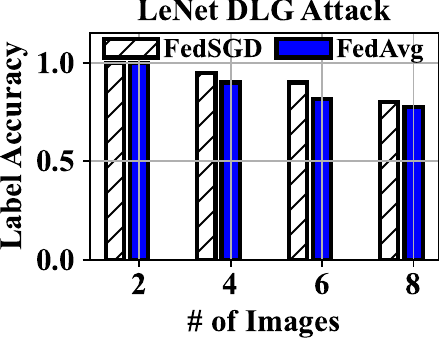}
		\label{fig:compare_fedsgd_fedavg_dlg_la}
	\end{subfigure}
	\hspace{-3mm}
	\begin{subfigure}[t]{0.48\linewidth}
	\centering
		\includegraphics[scale=0.48]{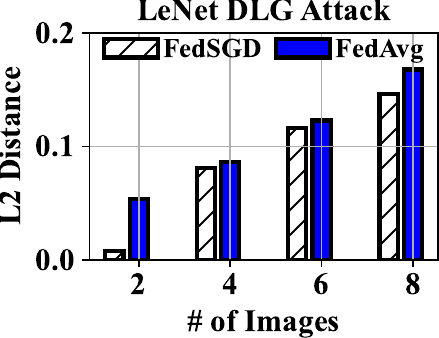}
		\label{fig:compare_fedsgd_fedavg_dlg_l2d}
	\end{subfigure}
	\vspace{-2mm}
	\caption{FedSGD vs. FedAvg under the gradient attacks.}
	\label{fig:comparing fedsgd and fedavg}
	\vspace{-2mm}
\end{figure}


In the gradient attack experiments, we calculate the gradients using parameters from two adjacent training steps. The attack results are evaluated using the label accuracy and L2 distance between attack results and real data. Data label is one of the outputs in the DLG attack. Thus we can directly compute its accuracy. In FC attack, we use a well-trained CNN model to determine whether the attack outputs and real images have the same label. To further compare FedSGD and FedAvg and demonstrate what affects the results of gradient attacks, we perform experiments in the following aspects: \textbf{1) Varying the number of images in FedSGD and FedAvg}: \Cref{fig:attack_vary_images} shows the attack results varying the number of images. With the increasing number of images, the label accuracy drops, and L2 distance rises in both FedSGD and FedAvg. Thus the gradient attacks are more difficult when the clients have more images; \textbf{2) Varying the number of epochs in FedAvg}: Since the number of local training epochs (\ie E) in FedSGD is fixed to one, we only present the attack results of FedAvg when changing the value of $E$, and \Cref{fig:attack_vary_epchos} shows the results. With the increasing number of training epochs, the L2 distance significantly rises in the DLG attack and slightly grows in the FC attack. We can conclude that the gradient attacks are more difficult when the clients perform more local training steps before uploading the parameters to the server. Moreover, the FC attack is less sensitive to $E$ compared with the DLG attack, and its attack error (\ie L2 distance) on ten images is still very low when the clients perform 100 rounds of local training; \textbf{3) FedSGD vs. FedAvg}: \Cref{fig:comparing fedsgd and fedavg} compares the attack results on FedSGD and FedAvg (B=1, E=20). In the FC attack, the label accuracy of FedSGD and FedAvg are very close, but the L2 distance of FedAvg is larger. In the DLG attack, FedAvg has lower attack label accuracy and larger L2 distance compared with FedSGD. In front of the FC attack, FedSGD has a better performance. In contrast, FedAvg has better performance under the DLG attack. One likely reason is that the FC attack is not easily influenced by the number of clients' local training steps; \textbf{4) FC attack vs. DLG attack}: The first subplot in \Cref{fig:attack_vary_images} compares the attack results of FC and DLG. These two methods' attack performances are almost the same when the client has five (or fewer) images. However, when the client has ten or more images, the FC attack significantly outperforms the DLG attack model (\ie FC attack has higher label accuracy and lower L2 distance).

To summarize, both FedSGD and FedAvg have privacy issues, and FedAvg performs better than FedSGD in resisting the FC and DLG attacks. When attacking MLP models, FC is more effective than DLG. Generally, gradient attacks are more difficult when the clients have more images or perform more local training epochs.


\vspace{-2mm}
\subsection{Effectiveness and Robustness Evaluation}

\begin{table}[h]
	\setlength{\tabcolsep}{0.05em}
	\renewcommand\arraystretch{0.7}
	\vspace{+5mm}
	\caption{Robustness and Effectiveness Benchmarks. All the experiments are repeated ten times, and the average values and standard error are reported. The MNIST dataset adopts the non-IID label setting while the other datasets adopt the non-IID feature settings.}
	\label{tab:robustness and efficacy}
	\vspace{-3mm}
	\centering
	\begin{threeparttable}
	\begin{tabular}{c|c|c|c|c|c|c|c|c}
		\toprule
		Dataset & IID & Local & Central & FedSGD & FedAvg & FedProx & FedSTC & FedOpt  \\
		\midrule
		mnist   & N & \multirow{3}{*}{\tabincell{c}{0.11319\\(0.013)}}  & \multirow{3}{*}{\tabincell{c}{0.98614\\(0.001)}} & \tabincell{c}{0.98390\\(0.001)} & \tabincell{c}{0.97843 \\ (0.006)} & \tabincell{c}{0.97874\\(0.003)} & \tabincell{c}{0.43416\\(0.065)} & \tabincell{c}{0.97679\\(0.003)} \\
		\cmidrule(l){1-2}
		\cmidrule(l){5-9}
		mnist   & Y &  &  & \tabincell{c}{0.98341\\(0.002)} & \tabincell{c}{0.98651\\(0.001)} & \tabincell{c}{0.98683\\(0.001)} & \tabincell{c}{0.94421\\(0.002)} &  \tabincell{c}{0.98351\\(0.001)} \\
		\midrule
		femnist & N & \multirow{3}{*}{\tabincell{c}{0.48231\\(0.056)}} & \multirow{3}{*}{\tabincell{c}{0.84961\\(0.002)}} & \tabincell{c}{0.80461\\(0.015)} & \tabincell{c}{0.81234\\(0.004)} & \tabincell{c}{0.81288\\(0.005)} & \tabincell{c}{0.44646\\(0.019)} & \tabincell{c}{0.80783\\(0.003)} \\
		\cmidrule(l){1-2}
		\cmidrule(l){5-9}
		femnist & Y &  &  & \tabincell{c}{0.81351\\(0.012)} & \tabincell{c}{0.83476\\(0.004)} & \tabincell{c}{0.83385\\(0.002)} & \tabincell{c}{0.49307\\(0.022)} & \tabincell{c}{0.83187\\(0.004)} \\
		\midrule
		celebA  & N & \multirow{3}{*}{\tabincell{c}{0.70307\\(0.007)}} & \multirow{3}{*}{\tabincell{c}{0.92400\\(0.005)}} & \tabincell{c}{0.91707\\(0.005)} & \tabincell{c}{0.90170\\(0.005)} & \tabincell{c}{0.90120\\(0.007)} & \tabincell{c}{0.66613\\(0.044)} & \tabincell{c}{0.89913\\(0.008)} \\
		\cmidrule(l){1-2}
		\cmidrule(l){5-9}
		celebA  & Y &  &  & \tabincell{c}{0.91867\\(0.006)} & \tabincell{c}{0.90267\\(0.012)} & \tabincell{c}{0.90210\\(0.011)} & \tabincell{c}{0.67840\\(0.058)} & \tabincell{c}{0.89957\\(0.011)} \\
		\midrule
		sent140 & N & \multirow{3}{*}{\tabincell{c}{0.74447\\(0.006)}} & \multirow{3}{*}{\tabincell{c}{0.79263\\(0.002)}} & \tabincell{c}{0.74131\\(0.006)} & \tabincell{c}{0.75578\\(0.003)} & \tabincell{c}{0.75626\\(0.003)} & \tabincell{c}{0.70954\\(0.006)} & \tabincell{c}{0.75263\\(0.004)} \\
		\cmidrule(l){1-2}
		\cmidrule(l){5-9}
		sent140 & Y &  &  & \tabincell{c}{0.74024\\(0.005)} & \tabincell{c}{0.76504\\(0.004)} & \tabincell{c}{0.75839\\(0.005)} & \tabincell{c}{0.69939\\(0.009)} & \tabincell{c}{0.74955\\(0.007)} \\
		\midrule
		Average & N & \multirow{3}{*}{\tabincell{c}{0.51076}} & \multirow{3}{*}{\tabincell{c}{0.88809}} & \tabincell{c}{0.86172} & \tabincell{c}{0.86206} & \tabincell{c}{0.86227} & \tabincell{c}{0.56407} & \tabincell{c}{0.85909} \\
		\cmidrule(l){1-2}
		\cmidrule(l){5-9}
		Average & Y &  &  & \tabincell{c}{0.86395} & \tabincell{c}{0.87224} & \tabincell{c}{0.87029} & \tabincell{c}{0.70376} & \tabincell{c}{0.86612} \\
		\bottomrule
	\end{tabular}
    \end{threeparttable}
	\vspace{-4mm}
\end{table}


\Cref{tab:robustness and efficacy} shows the effectiveness and robustness evaluation of five FL mechanisms\footnote{Currently, we mainly focus on evaluating non-IID data problems. The evaluation of system uncertainties will be done in the future.}. SecAgg and HEAgg are omitted in the effectiveness and robustness evaluation since they do not impact the model accuracy. Theoretically, they have the same performance as FedAvg, which is used as the base for implementing SecAgg and HEAgg. 

\parab{Effectiveness Evaluation:} We evaluate the FL algorithm's effectiveness by comparing their model accuracy with local and centralized training. We choose the FL algorithm's model accuracy under the non-IID setting in the comparison because real-world applications typically have non-IID data. The results in \Cref{tab:robustness and efficacy} show that: 1) Most of the FL algorithms significantly improve the model accuracy compared with local training which shows the effectiveness of federated learning; 2) All the FL algorithms have accuracy drop compared with centralized training showing that existing FL algorithms could be improved regarding the effectiveness.

\parab{Robustness Evaluation:} Based on the results in \Cref{tab:robustness and efficacy}, we have the following observation regarding the robustness evaluation: 1) All the tested FL algorithms' model effectiveness increases when changing the data setting from non-IID to IID; 2) FedSTC has the most significant performance disparity between non-IID and IID settings, showing that gradients compression methods are easier to be affected by non-IID data; 3) FedProx has the best model performance under the non-IID setting; 

\vspace{-2mm}
\subsection{Efficiency Evaluation}
\vspace{-2mm}

\begin{figure}[!h]
	\centering
	\includegraphics[width=0.5\textwidth]{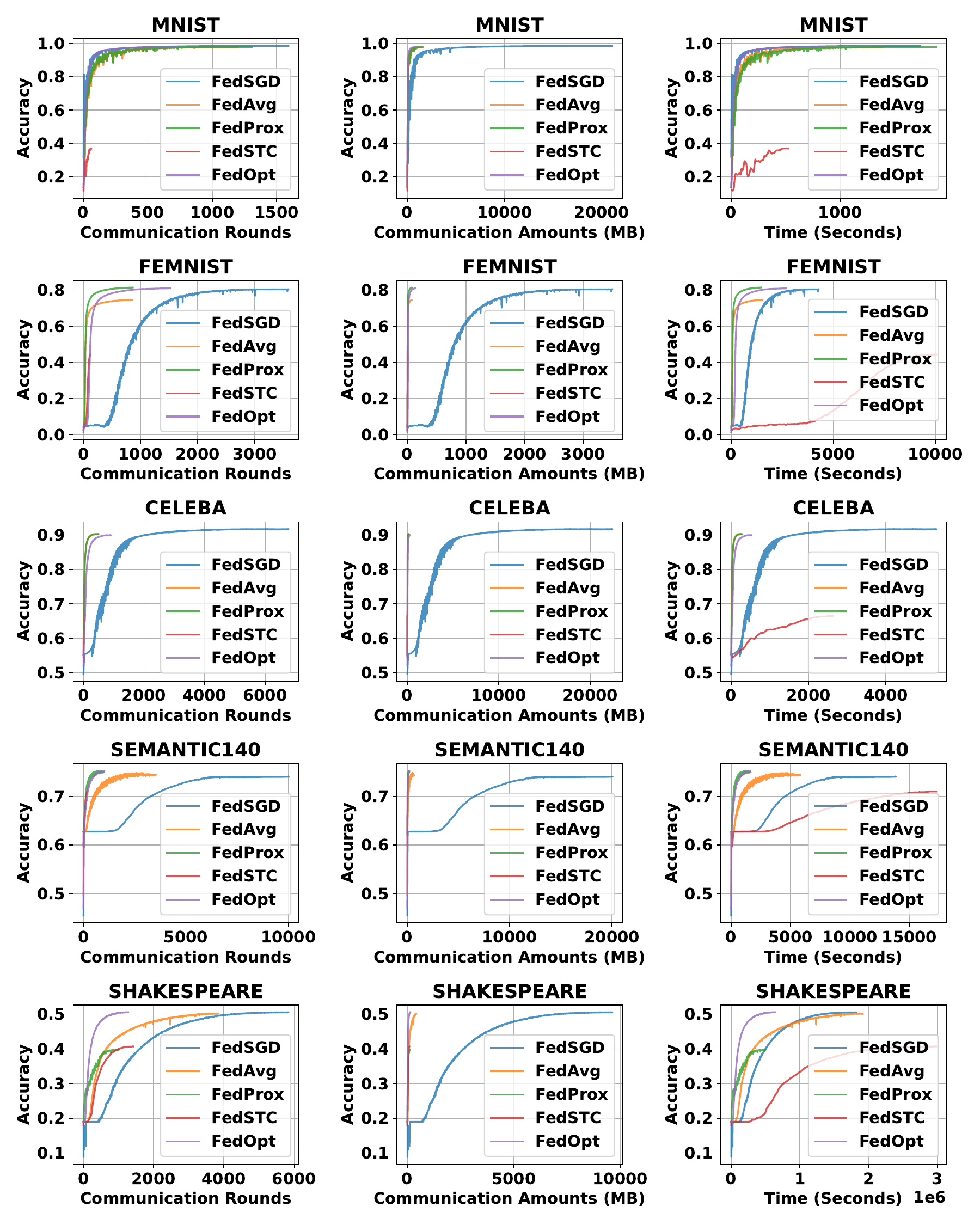}
	\vspace{-7mm}
	\caption{Efficiency evaluation. The lines stop when reaching the best validation loss. The communication amount are the sum of sending and receiving messages of individual client.}
	\label{fig:to_acc_lines}
\end{figure}

\begin{table}[h]
	\setlength{\tabcolsep}{0.3em}
	\renewcommand\arraystretch{0.1}
	\vspace{+6mm}
	\caption{The communication round (CR), communication amount (CA), and time consumption (hours) required for the FL algorithms to converge. (FedSTC is removed from the table because it fails to converge, and FedProx fails to converge on the Shakespeare dataset.)}
	\vspace{-2mm}
	\label{tab:efficiency table}
	\centering
	\begin{threeparttable}
	\begin{tabular}{c|c|c|c|c|c}
		\toprule
		Dataset & Metrics & FedSGD & FedAvg & FedProx & FedOpt  \\
		\midrule
		\multirow{5}{*}{MNIST} & \tabincell{c}{CR} & \tabincell{c}{1396} & \tabincell{c}{839} & \tabincell{c}{897} & \tabincell{c}{609} \\
		\cmidrule(l){2-6}
		& \tabincell{c}{CA} & \tabincell{c}{1.76TB} & \tabincell{c}{99.97GB} & \tabincell{c}{106.90GB} & \tabincell{c}{72.59GB} \\
		\cmidrule(l){2-6}
		& \tabincell{c}{Time} & \tabincell{c}{0.39} & \tabincell{c}{0.27} & \tabincell{c}{0.35} & \tabincell{c}{0.19} \\
		\midrule
		\multirow{5}{*}{FEMNIST} & \tabincell{c}{CR} & \tabincell{c}{2637} & \tabincell{c}{791} & \tabincell{c}{806} & \tabincell{c}{1261} \\
		\cmidrule(l){2-6}
		& \tabincell{c}{CA} & \tabincell{c}{2.45TB} & \tabincell{c}{69.24GB} & \tabincell{c}{70.50GB} & \tabincell{c}{110.38GB} \\
		\cmidrule(l){2-6}
		& \tabincell{c}{Time} & \tabincell{c}{0.73} & \tabincell{c}{0.39} & \tabincell{c}{0.38} & \tabincell{c}{0.62} \\
		\midrule
		\multirow{5}{*}{CelebA} & \tabincell{c}{CR}  & \tabincell{c}{6323} & \tabincell{c}{448} & \tabincell{c}{439} & \tabincell{c}{736} \\
		\cmidrule(l){2-6}
		& \tabincell{c}{CA} & \tabincell{c}{19.95TB} & \tabincell{c}{133.41GB} & \tabincell{c}{130.78GB} & \tabincell{c}{219.19GB} \\
		\cmidrule(l){2-6}
		& \tabincell{c}{Time} & \tabincell{c}{1.05} & \tabincell{c}{0.06} & \tabincell{c}{0.07} & \tabincell{c}{0.11} \\
		\midrule
		\multirow{5}{*}{Sent140} & \tabincell{c}{CR} & \tabincell{c}{6386} & \tabincell{c}{3013} & \tabincell{c}{899} & \tabincell{c}{964} \\
		\cmidrule(l){2-6}
		& \tabincell{c}{CA} & \tabincell{c}{12.19TB} & \tabincell{c}{542.78GB} & \tabincell{c}{163.05GB} & \tabincell{c}{174.67GB} \\
		\cmidrule(l){2-6}
		& \tabincell{c}{Time} & \tabincell{c}{2.13} & \tabincell{c}{1.36} & \tabincell{c}{0.40} & \tabincell{c}{0.41} \\
		\midrule
		\multirow{5}{*}{\tabincell{c}{Shakespeare}} & \tabincell{c}{CR} & \tabincell{c}{4835} & \tabincell{c}{3485} & \tabincell{c}{$\backslash$} & \tabincell{c}{1094} \\
		\cmidrule(l){2-6}
		& \tabincell{c}{CA} & \tabincell{c}{0.76TB} & \tabincell{c}{37.90GB} & \tabincell{c}{$\backslash$} & \tabincell{c}{11.88GB} \\
		\cmidrule(l){2-6}
		& \tabincell{c}{Time} & \tabincell{c}{417.7} & \tabincell{c}{482.60} & \tabincell{c}{$\backslash$} & \tabincell{c}{153.74} \\
		\bottomrule
	\end{tabular}
    \end{threeparttable}
	\vspace{-5mm}
\end{table}

\Cref{fig:to_acc_lines} shows the plot of communication round (CR) to accuracy, communication amount (CA) to accuracy, and time to accuracy. To make the comparison easier, we also report the final required CR, CA, and time to converge of each FL algorithm in \Cref{tab:efficiency table}. Based on the efficiency evaluation results, we have the following observations: 1) FedSGD requires significantly more communication rounds, communication amount, and time to converge compared with the other methods; 2) On the Shakespeare dataset, although FedAvg requires fewer communication rounds than FedSGD to converge, it has higher time consumption than FedSGD. The reason is that the clients in the Shakespeare dataset have more local training samples and FedAvg reduces the communication rounds by increasing the clients' local training passes in each round of training, which significantly increases the clients' local training time. Such issue happens when the clients have a large number of local training samples or the machine learning model is complex (\eg, deep learning models); 3) FedOpt significantly outperforms the other solutions on the Shakespeare dataset, and it shows that the federated optimizer solution better adapts to large datasets and complex models.

\begin{figure}[!h]
	\centering
	\includegraphics[scale=0.5]{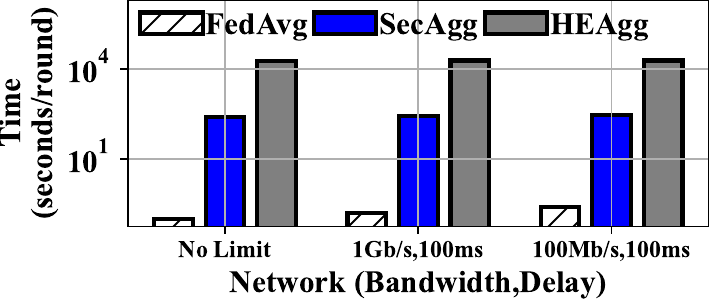}
	\vspace{-2mm}
	\caption{Comparing efficiency of FedAvg, SecAgg, and HEAgg.}
	\label{fig:network_secagg_paillier}
\end{figure}

\parab{Quantifying the efficiency overhead of SecAgg and HEAgg:} \Cref{fig:network_secagg_paillier} shows the time consumption of SecAgg, HEAgg, and FedAvg under different network bandwidth and latency. SecAgg is approximately two orders of magnitude slower than FedAvg, and HEAgg is approximately two orders of magnitude slower than SecAgg. The results show that although SecAgg and HEAgg have higher security levels than FedAvg but bring significant efficiency overhead.

\vspace{-4mm}
\subsection{Visualization through Radar Charts}

\begin{figure*}[h!]
	\center
	\begin{subfigure}[t]{0.14\textwidth}
		\includegraphics[width=\textwidth]{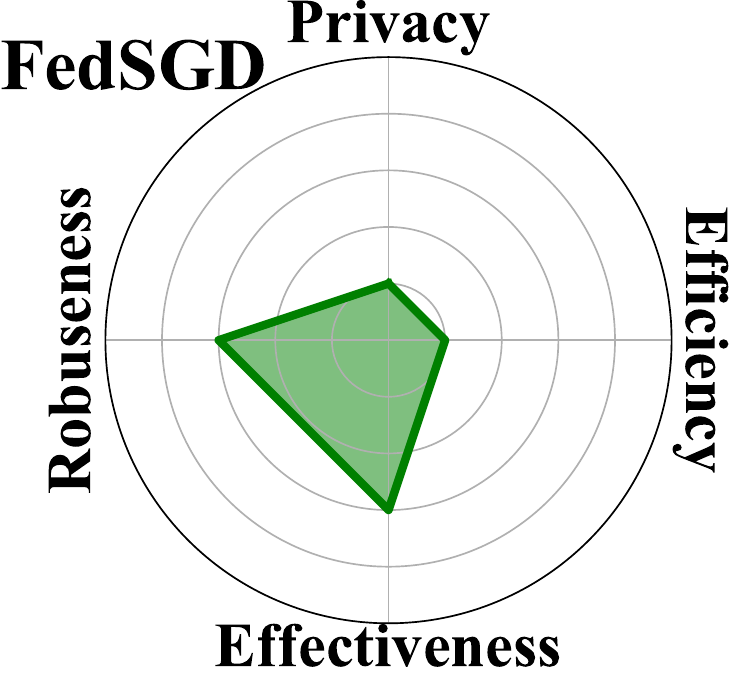}
	\end{subfigure}
	\hspace{-2mm}
	\begin{subfigure}[t]{0.14\textwidth}
		\includegraphics[width=\textwidth]{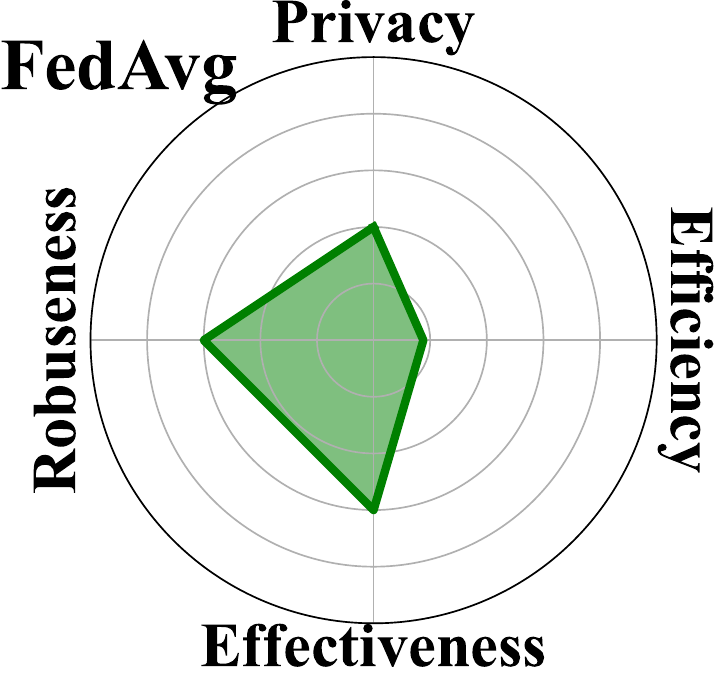}
	\end{subfigure}
	\hspace{-2mm}
	\begin{subfigure}[t]{0.14\textwidth}
		\includegraphics[width=\textwidth]{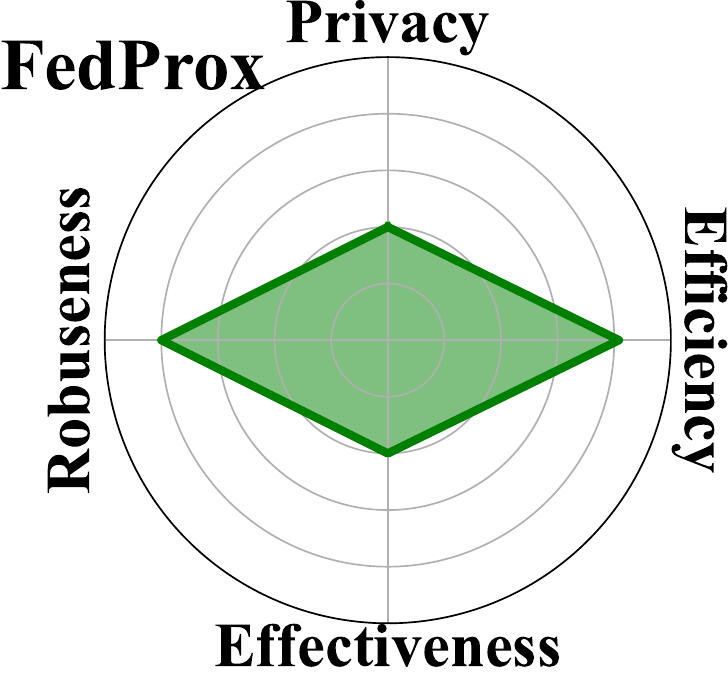}
	\end{subfigure}
	\hspace{-2mm}
	\begin{subfigure}[t]{0.14\textwidth}
		\includegraphics[width=\textwidth]{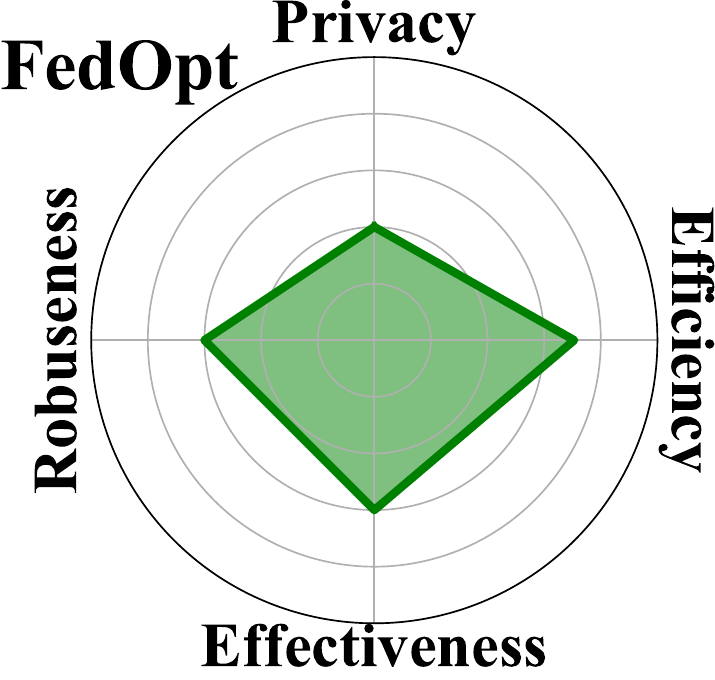}
	\end{subfigure}
	\hspace{-2mm}
	\begin{subfigure}[t]{0.14\textwidth}
		\includegraphics[width=\textwidth]{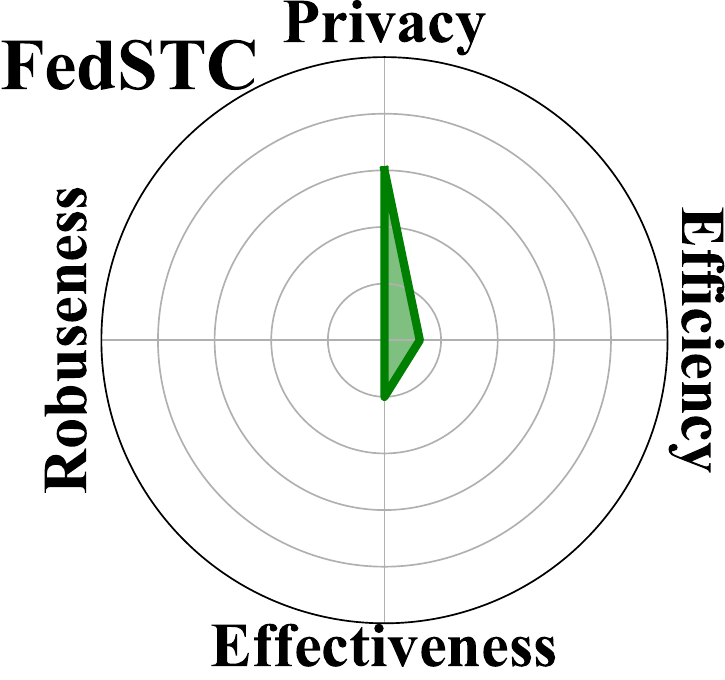}
	\end{subfigure}
	\hspace{-2mm}
	\begin{subfigure}[t]{0.14\textwidth}
		\includegraphics[width=\textwidth]{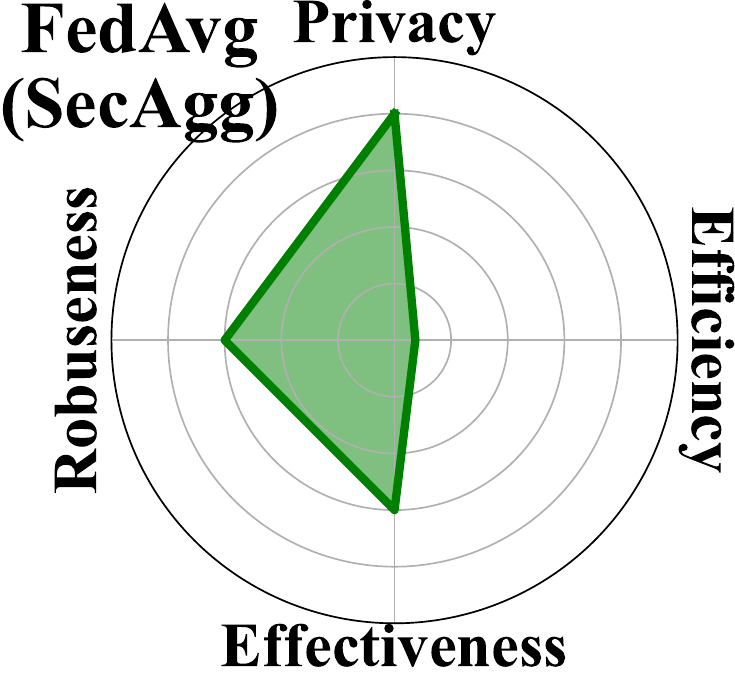}
	\end{subfigure}
	\hspace{-2mm}
	\begin{subfigure}[t]{0.14\textwidth}
		\includegraphics[width=\textwidth]{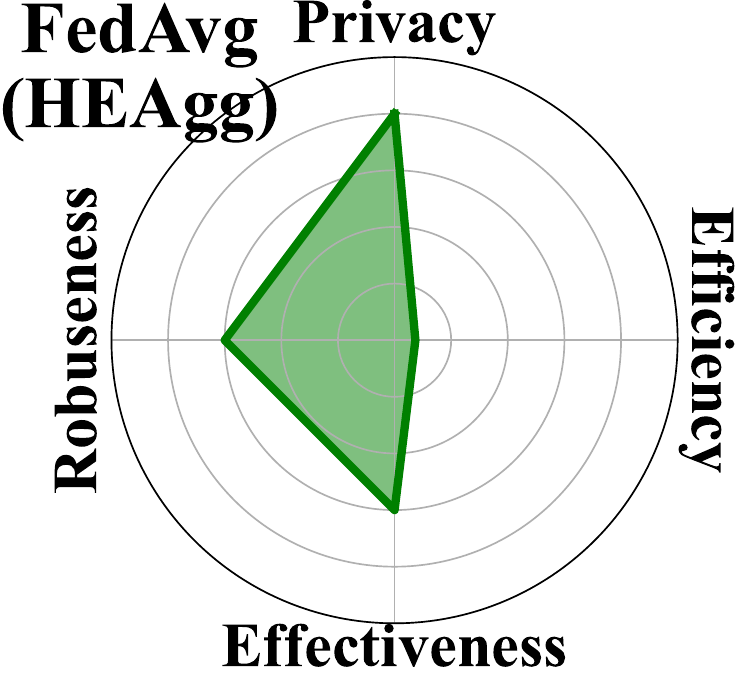}
	\end{subfigure}
	\vspace{-2mm}
	\caption{Visualizing the benchmarking results through radar charts.}
	\label{fig:rader_charts}
	\vspace{-2mm}
\end{figure*}

To give a more straightforward illustration and clear comparison, we propose visualizing the evaluation results through radar charts, which are presented in \Cref{fig:rader_charts}. The scores presented in the radar chart are computed by combining both quantitative and qualitative measurements. Due to the space limitation, we put the detailed computation method in our online documentation\footnote{\url{https://fedeval.readthedocs.io/en/latest/benchmark/benchmark.html}}.

\vspace{-4mm}
\subsection{Suitable Scenarios of the Evaluated FL Algorithms}

To provide practical guidance for applying these FL algorithms in real-world applications, we analyze the suitable scenarios for each algorithm based on the evaluation results.

\begin{icompact}
	\item FedSGD/FedAvg/FedProx: These three algorithms are highly correlated. FedProx improves the performance under heterogeneous data compared with FedAvg by setting a regularization such that the distance between the local and global model is not too large. And by setting the regularization term to zero, FedProx is equivalent to FedAvg. FedAvg improves the communication efficiency compared with FedSGD by increasing the local training rounds and only selects a small part of the clients (\eg, 10\%) in each round of training. And by setting the local training round to one and letting all the clients participate in the training, FedAvg is equivalent to FedSGD. Thus, in real-world applications, users could smoothly transfer between these three algorithms by adjusting the ratio of the participated clients, the clients' local training rounds, and the regularization term such that the algorithm gets the best performance regarding robustness and efficiency.
	\item FedOpt is suitable for scenarios with large models and datasets to accelerate the training. For example, in our evaluation, FedOpt has the best efficiency on the Shakespeare dataset, which has the most training samples.
	\item FedSTC is suitable for scenarios with very low networking bandwidth since it has a significantly smaller communication amount than other algorithms. However, we suggest fine-tuning the sparsity carefully since our evaluation shows that high sparsity (\eg, 1\%) may impact the model's effectiveness.
\end{icompact}

\section{Insights and Identified Future Research Directions} \label{sec:insights}

Based on our investigation and benchmarking studies, we summarize the following insights and future research directions for researchers in the FL area:

\parab{Insight 1: Privacy is overlooked in existing FL mechanism designs.} Privacy protection is the foundation of FL. However, based on our investigations, most FL algorithms belong to level-1 security in which the adversaries learn the individual intermediate results (\ie, gradients) in plaintext. According to our privacy evaluation on two representative work in level-1 security, \ie FedSGD and FedAvg, directly exchanging individual gradients is not secure. We strongly suggest that new FL studies should analyze and perform evaluations regarding privacy preservation.

\parab{Insight 2: A comprehensive evaluation is needed for FL studies.} During the benchmarking study, we have found the following problems caused by the incomprehensive evaluation metrics: 1) FedAvg reduces the communication round by increasing clients' local training workload, however, it brings larger overall time consumption on large-scale datasets (\eg, Shakespeare) because the edge devices usually have limited computing resources; 2) FedSTC adopts the gradients compression method to reduce the communication amount, however, it brings a significant effectiveness decrease, especially under the non-IID settings; 3) The SecAgg and HEAgg provide better privacy preservation through adopting encryption techniques (\eg, key exchange, public-key encryption, homomorphic encryption, \etc.). However, strictly following these encryption-based protocols brings extensive computation and communication overhead, making them not practical for real-world applications with high-efficiency requirements;

\parab{Insight 3: Consistent efficiency metrics are needed.} During the benchmarking study on efficiency, we have found that the FL algorithms usually have inconsistent performance regarding communication amount, communication round, and time consumption. Given two FL mechanisms, \eg model A and model B, it is hard to compare them if A is better than B in communication amount but worse than B in communication rounds. Thus, compatible efficiency metrics are needed. For example, \cite{flower} has considered using carbon emission as the evaluation metric, which could be widely accepted because FL should be environmentally friendly.

\parab{Future Directions 1: Level-3 safety with high efficiency.} Privacy preservation is one of the most fundamental problems in FL and needs to be properly handled. According to our categorization in \Cref{tab:security_level}, level-3 safety provides the best protection in FL, and the server learns nothing about individual clients. However, existing work achieves level-3 safety through fully homomorphic encryption (FHE), which brings large computation overhead, making it not practical in real-world applications. Thus designing an FL method that simultaneously achieves level-3 safety and high efficiency can significantly benefit real-world applications. Existing work \cite{FedSVD} has explored using random masks (\ie, non-FHE solutions) to achieve level-3 safety and keep high efficiency. However, it is still limited to specific FL tasks (\eg singular vector decomposition), and its extensibility to general ML model training still needs to be further studied.

\parab{Future Directions 2: Effective FL algorithm with very low communication amount overhead.} Communication amount is an essential metric in FL because 1) Most edge devices participating in the FL training have limited networking bandwidth; 2) Transferring a large amount of data is not permitted in some scenarios where the participants have high privacy protection requirements, \eg, banks. And the problem of reducing communication amount is still under-investigated in the FL literature. The evaluation results in \Cref{tab:efficiency table} show that all the benchmarked FL algorithms have a large amount of communication, and the gradients-compression-based solution (\ie, FedSTC) suffers from the issue of effectiveness drop under the non-IID data settings. Thus proposing a new FL algorithm that can finish the federated training using very few amounts of communication is a good future research direction and will have many applications in the real world.

\vspace{-3mm}
\section{Related Work} \label{sec:related_work}
\vspace{-2mm}

\parab{FL Platforms:} Since the idea of FL was proposed, many platforms and tools have been developed to help implement new FL algorithms in both research and industry areas, \eg, FATE~\cite{FATE}, FedScale~\cite{fedscale}, FedML~\cite{he2020fedml}, \etc. These FL platforms focus on providing easy-to-use, scalable, and secure FL services or APIs to implement new FL algorithms. Our study differs from FL platform work in the following aspects: 1) Although existing platforms largely cover the evaluation aspects or the evaluation aspects could be quickly added if not covered, the evaluation metrics are not well organized, and the evaluation processes are not standardized. Our study focuses on properly organizing these metrics, proposing the taxonomy model, and standardizing the evaluation such that we can evaluate and analyze FL algorithms from a new perspective. We believe that \fedeval, especially the \fedeval-Core evaluation taxonomy model, could inspire such FL platforms to design and implement a standard holistic evaluation procedure. 2) From the benchmark perspective, the evaluation results reported in the FL platform papers~\cite{fedscale} mainly focus on showing the platform's usability, while our evaluation focuses on holistically evaluating the ability of FL algorithms from different aspects and obtaining practical guidelines. Meanwhile, \fedeval could also be generalized from evaluating FL algorithm to evaluating these platforms, \eg, the same FL algorithm may have different performance under different platforms due to the different backends and implementations, and we take this as our future work.

\begin{table}[h!]
	\centering
	\renewcommand\arraystretch{1.0}
	\setlength{\tabcolsep}{0.5em}
	\vspace{+2mm}
	\caption{Comparing \fedeval with other FL benchmarks.}
	\label{tab:FLBenchmark}
	\vspace{-2mm}
	\begin{tabular}{c|c|c|c|c|c}
		\toprule
		\tabincell{c}{Evaluation\\Aspects} & \tabincell{c}{LEAF \\ \cite{caldas2018leaf}} & \tabincell{c}{ORAF \\ \cite{hu2020oarf}} & \tabincell{c}{Nilsson \\ et al.~\cite{DBLP:conf/middleware/NilssonSUGJ18}} & \tabincell{c}{EFFL \\ \cite{liu2020evaluation}} & \fedeval \\ \midrule
		Privacy &  & \LEFTcircle && & \CIRCLE \\ \midrule
		Non-IID Feature & \CIRCLE &  && \CIRCLE & \CIRCLE \\ \midrule
		Non-IID Label & \CIRCLE & \CIRCLE && \CIRCLE & \CIRCLE \\ \midrule
		HeteroSys &&&& & \CIRCLE \\ \midrule
		CommRound && \CIRCLE & \CIRCLE & & \CIRCLE \\ \midrule
		CommAmount && \CIRCLE && & \CIRCLE \\ \midrule
		Time &&&&& \CIRCLE \\ \midrule
		Effectiveness & \LEFTcircle & \LEFTcircle & \LEFTcircle & & \CIRCLE \\
	    \bottomrule
	\end{tabular}
	\vspace{-2mm}
\end{table}

\parab{FL Benchmarks:} There are also many other studies that point out the non-comparable FL evaluation problems and try to give a standardized evaluation. \citet{caldas2018leaf} compared FedSGD and FedAvg regarding the non-IID issue on the six datasets. \citet{hu2020oarf} proposed a benchmark suite that evaluates FL methods in model utility, communication cost, privacy loss, and encryption overhead. \citet{DBLP:conf/middleware/NilssonSUGJ18} benchmarked three federated learning models using one dataset, and they mainly focused on comparing the effectiveness of FL models with the centralized training. Compared with existing FL benchmark studies, as illustrated in \Cref{tab:FLBenchmark}, our work proposes a holistic evaluation taxonomy model that provides more thorough and convincing FL benchmark results. Additionally, we also provide an evaluation platform that can further standardize and simplify the evaluation process.

\vspace{-4mm}
\section{Conclusion and Future Work}

In this paper, we propose an evaluation framework for FL, called \fedeval, which is pressingly needed to unreservedly present the advantages and disadvantages of FL algorithms and compare existing and future works from a common standpoint. Based on \fedeval, we have performed a benchmarking study between seven well-known FL mechanisms: FedSGD, FedAvg, FedProx, FedSTC, FedOpt, SecAgg, and HEAgg. We also have drawn a set of take-away insights and future research directions, which are very helpful for researchers in the FL area.

In the future, we will try to improve \fedeval in the following directions:

\begin{icompact}
	\item Explore more scenarios of attack and defenses regarding privacy and security. Apart from the gradients attacks evaluated in this paper, there are also many other scenarios of attacks and defenses in FL, \eg, backdoor attacks \cite{bagdasaryan2018backdoor}, model inversion attacks \cite{hidano2018model,fredrikson2015model}, membership inference attacks \cite{zhang2020gan}, \etc. Exploring these attacks in \fedeval will help provide better privacy and security evaluations.
	\item Extending \fedeval to evaluate different FL platforms. In the future, we will extend \fedeval to evaluate different FL platforms. Briefly, the same FL algorithm might be implemented differently in different platforms, \eg, different backends are used and variations to the FL algorithm might be made to make the implementation easier. Thus, evaluating different FL platforms under the same FL algorithm is necessary to verify and compare their correctness, efficiency, privacy preservation, \etc.
\end{icompact}

\vspace{-4mm}


%

%

\vspace{-1mm}
\section*{Acknowledgments}
The work is supported by the Key-Area Research and Development Program of Guangdong Province (2021B0101400001), the NSFC Grant no. 61972008, the Hong Kong RGC TRS T41-603/20R, the National Key Research and Development Program of China under Grant No.2018AAA0101100.
\vspace{-3mm}


\ifCLASSOPTIONcaptionsoff
  \newpage
\fi



\bibliographystyle{IEEEtranN}
\bibliography{sections/ref.bib}

\begin{thebibliography}{81}
\providecommand{\natexlab}[1]{#1}
\providecommand{\url}[1]{#1}
\csname url@samestyle\endcsname
\providecommand{\newblock}{\relax}
\providecommand{\bibinfo}[2]{#2}
\providecommand{\BIBentrySTDinterwordspacing}{\spaceskip=0pt\relax}
\providecommand{\BIBentryALTinterwordstretchfactor}{4}
\providecommand{\BIBentryALTinterwordspacing}{\spaceskip=\fontdimen2\font plus
\BIBentryALTinterwordstretchfactor\fontdimen3\font minus
  \fontdimen4\font\relax}
\providecommand{\BIBforeignlanguage}[2]{{%
\expandafter\ifx\csname l@#1\endcsname\relax
\typeout{** WARNING: IEEEtranN.bst: No hyphenation pattern has been}%
\typeout{** loaded for the language `#1'. Using the pattern for}%
\typeout{** the default language instead.}%
\else
\language=\csname l@#1\endcsname
\fi
#2}}
\providecommand{\BIBdecl}{\relax}
\BIBdecl

\bibitem[McMahan et~al.(2017)McMahan, Moore, Ramage, Hampson, and
  y~Arcas]{mcmahan2016communication}
B.~McMahan, E.~Moore, D.~Ramage, S.~Hampson, and B.~A. y~Arcas,
  ``Communication-efficient learning of deep networks from decentralized
  data,'' in \emph{{AISTATS}}, 2017.

\bibitem[Li et~al.(2020{\natexlab{a}})Li, Sahu, Zaheer, Sanjabi, Talwalkar, and
  Smith]{FedProx}
T.~Li, A.~K. Sahu, M.~Zaheer, M.~Sanjabi, A.~Talwalkar, and V.~Smith,
  ``Federated optimization in heterogeneous networks,'' \emph{MLSys}, 2020.

\bibitem[Reddi et~al.(2021)Reddi, Charles, Zaheer, Garrett, Rush,
  Kone{\v{c}}n{\'y}, Kumar, and McMahan]{FedOpt}
S.~J. Reddi, Z.~Charles, M.~Zaheer, Z.~Garrett, K.~Rush, J.~Kone{\v{c}}n{\'y},
  S.~Kumar, and H.~B. McMahan, ``Adaptive federated optimization,'' in
  \emph{{ICLR}}, 2021.

\bibitem[Sattler et~al.(2020)Sattler, Wiedemann, M{\"{u}}ller, and
  Samek]{FedSTC}
F.~Sattler, S.~Wiedemann, K.~M{\"{u}}ller, and W.~Samek, ``Robust and
  communication-efficient federated learning from non-i.i.d. data,''
  \emph{{IEEE} Trans. Neural Networks Learn. Syst.}, 2020.

\bibitem[Bonawitz et~al.(2017{\natexlab{a}})Bonawitz, Ivanov, Kreuter,
  Marcedone, McMahan, Patel, Ramage, Segal, and Seth]{SecureAggregation}
K.~A. Bonawitz, V.~Ivanov, B.~Kreuter, A.~Marcedone, H.~B. McMahan, S.~Patel,
  D.~Ramage, A.~Segal, and K.~Seth, ``Practical secure aggregation for
  privacy-preserving machine learning,'' in \emph{{CCS}}, 2017.

\bibitem[Aono et~al.(2017)Aono, Hayashi, Wang, Moriai, et~al.]{aono2017privacy}
Y.~Aono, T.~Hayashi, L.~Wang, S.~Moriai \emph{et~al.}, ``Privacy-preserving
  deep learning via additively homomorphic encryption,'' \emph{{IEEE} Trans.
  Inf. Forensics Secur.}, 2017.

\bibitem[Chai et~al.(2020)Chai, Wang, Chen, and Yang]{chai2020secure}
D.~Chai, L.~Wang, K.~Chen, and Q.~Yang, ``Secure federated matrix
  factorization,'' \emph{IEEE Intelligent Systems}, 2020.

\bibitem[Li et~al.(2020{\natexlab{b}})Li, Kovalev, Qian, and
  Richtarik]{pmlr-v119-li20g}
Z.~Li, D.~Kovalev, X.~Qian, and P.~Richtarik, ``Acceleration for compressed
  gradient descent in distributed and federated optimization,'' in
  \emph{{ICML}}, 2020.

\bibitem[Yang et~al.(2021)Yang, Fang, and Liu]{DBLP:conf/iclr/YangFL21}
H.~Yang, M.~Fang, and J.~Liu, ``Achieving linear speedup with partial worker
  participation in non-iid federated learning,'' in \emph{{ICLR}}, 2021.

\bibitem[Lai et~al.(2021{\natexlab{a}})Lai, Zhu, Madhyastha, and
  Chowdhury]{Oort}
F.~Lai, X.~Zhu, H.~V. Madhyastha, and M.~Chowdhury, ``Oort: Efficient federated
  learning via guided participant selection,'' in \emph{{OSDI}}, 2021.

\bibitem[Sav et~al.(2021{\natexlab{a}})Sav, Pyrgelis, Troncoso{-}Pastoriza,
  Froelicher, Bossuat, Sousa, and Hubaux]{POSEIDON}
S.~Sav, A.~Pyrgelis, J.~R. Troncoso{-}Pastoriza, D.~Froelicher, J.~Bossuat,
  J.~S. Sousa, and J.~Hubaux, ``{POSEIDON:} privacy-preserving federated neural
  network learning,'' in \emph{{NDSS}}, 2021.

\bibitem[Shamsian et~al.(2021)Shamsian, Navon, Fetaya, and
  Chechik]{DBLP:conf/icml/ShamsianNFC21}
A.~Shamsian, A.~Navon, E.~Fetaya, and G.~Chechik, ``Personalized federated
  learning using hypernetworks,'' in \emph{{ICML}}, 2021.

\bibitem[Acar et~al.(2021{\natexlab{a}})Acar, Zhao, Zhu, Navarro, Mattina,
  Whatmough, and Saligrama]{DBLP:conf/icml/AcarZZNMWS21}
D.~A.~E. Acar, Y.~Zhao, R.~Zhu, R.~M. Navarro, M.~Mattina, P.~N. Whatmough, and
  V.~Saligrama, ``Debiasing model updates for improving personalized federated
  training,'' in \emph{{ICML}}, 2021.

\bibitem[Collins et~al.(2021)Collins, Hassani, Mokhtari, and
  Shakkottai]{DBLP:conf/icml/CollinsHMS21}
L.~Collins, H.~Hassani, A.~Mokhtari, and S.~Shakkottai, ``Exploiting shared
  representations for personalized federated learning,'' in \emph{{ICML}},
  2021.

\bibitem[Hanzely et~al.(2020)Hanzely, Hanzely, Horv{\'{a}}th, and
  Richt{\'{a}}rik]{DBLP:conf/nips/HanzelyHHR20}
F.~Hanzely, S.~Hanzely, S.~Horv{\'{a}}th, and P.~Richt{\'{a}}rik, ``Lower
  bounds and optimal algorithms for personalized federated learning,'' in
  \emph{NeurIPS}, 2020.

\bibitem[Diao et~al.(2021)Diao, Ding, and Tarokh]{DBLP:conf/iclr/Diao0T21}
E.~Diao, J.~Ding, and V.~Tarokh, ``Heterofl: Computation and communication
  efficient federated learning for heterogeneous clients,'' in \emph{{ICLR}},
  2021.

\bibitem[Dennis et~al.(2021)Dennis, Li, and Smith]{DBLP:conf/icml/Dennis0S21}
D.~K. Dennis, T.~Li, and V.~Smith, ``Heterogeneity for the win: One-shot
  federated clustering,'' in \emph{{ICML}}, 2021.

\bibitem[Zhao et~al.(2018)Zhao, Li, Lai, Suda, Civin, and
  Chandra]{zhao2018federated}
Y.~Zhao, M.~Li, L.~Lai, N.~Suda, D.~Civin, and V.~Chandra, ``Federated learning
  with non-iid data,'' \emph{arXiv}, 2018.

\bibitem[Zhu et~al.(2019)Zhu, Liu, and Han]{zhu2019deep}
L.~Zhu, Z.~Liu, and S.~Han, ``Deep leakage from gradients,'' in
  \emph{{NeurIPS}}, 2019.

\bibitem[Caldas et~al.(2018{\natexlab{a}})Caldas, Wu, Li, Kone{\v{c}}n{\`y},
  McMahan, Smith, and Talwalkar]{caldas2018leaf}
S.~Caldas, P.~Wu, T.~Li, J.~Kone{\v{c}}n{\`y}, H.~B. McMahan, V.~Smith, and
  A.~Talwalkar, ``Leaf: A benchmark for federated settings,'' \emph{arXiv},
  2018.

\bibitem[Hu et~al.(2020)Hu, Li, Liu, Li, Wu, and He]{hu2020oarf}
S.~Hu, Y.~Li, X.~Liu, Q.~Li, Z.~Wu, and B.~He, ``The oarf benchmark suite:
  Characterization and implications for federated learning systems,''
  \emph{arXiv}, 2020.

\bibitem[Nilsson et~al.(2018)Nilsson, Smith, Ulm, Gustavsson, and
  Jirstrand]{DBLP:conf/middleware/NilssonSUGJ18}
A.~Nilsson, S.~Smith, G.~Ulm, E.~Gustavsson, and M.~Jirstrand, ``A performance
  evaluation of federated learning algorithms,'' in \emph{DIDL@Middleware},
  2018.

\bibitem[Liu et~al.(2020)Liu, Zhang, Xiao, and Wu]{liu2020evaluation}
L.~Liu, F.~Zhang, J.~Xiao, and C.~Wu, ``Evaluation framework for large-scale
  federated learning,'' \emph{arXiv}, 2020.

\bibitem[Liu et~al.(2021)Liu, Fan, Chen, Xu, and Yang]{FATE}
Y.~Liu, T.~Fan, T.~Chen, Q.~Xu, and Q.~Yang, ``Fate: An industrial grade
  platform for collaborative learning with data protection,'' \emph{{JMLR}},
  2021.

\bibitem[Lai et~al.(2022)Lai, Dai, Singapuram, Liu, Zhu, Madhyastha, and
  Chowdhury]{fedscale}
F.~Lai, Y.~Dai, S.~S. Singapuram, J.~Liu, X.~Zhu, H.~V. Madhyastha, and
  M.~Chowdhury, ``{FedScale}: Benchmarking model and system performance of
  federated learning at scale,'' in \emph{{ICML}}, 2022.

\bibitem[Zhang et~al.(2022)Zhang, Gu, Fan, Chen, and Yang]{NFL}
X.~Zhang, H.~Gu, L.~Fan, K.~Chen, and Q.~Yang, ``No free lunch theorem for
  security and utility in federated learning,'' \emph{arXiv preprint
  arXiv:2203.05816}, 2022.

\bibitem[Ma et~al.(2022)Ma, Zhu, Lin, Chen, and Qin]{ma2022state}
X.~Ma, J.~Zhu, Z.~Lin, S.~Chen, and Y.~Qin, ``A state-of-the-art survey on
  solving non-iid data in federated learning,'' \emph{Future Generation
  Computer Systems}, 2022.

\bibitem[Smith et~al.(2017{\natexlab{a}})Smith, Chiang, Sanjabi, and
  Talwalkar]{smith2017federated}
V.~Smith, C.-K. Chiang, M.~Sanjabi, and A.~S. Talwalkar, ``Federated multi-task
  learning,'' in \emph{{NeurIPS}}, 2017, pp. 4424--4434.

\bibitem[Yang et~al.(2019)Yang, Liu, Chen, and Tong]{yang2019federated}
Q.~Yang, Y.~Liu, T.~Chen, and Y.~Tong, ``Federated machine learning: Concept
  and applications,'' \emph{ACM TIST}, 2019.

\bibitem[Li et~al.(2020{\natexlab{c}})Li, Sanjabi, Beirami, and
  Smith]{li2019fair}
T.~Li, M.~Sanjabi, A.~Beirami, and V.~Smith, ``Fair resource allocation in
  federated learning,'' in \emph{{ICLR}}, 2020.

\bibitem[Smith et~al.(2017{\natexlab{b}})Smith, Chiang, Sanjabi, and
  Talwalkar]{DBLP:conf/nips/SmithCST17}
V.~Smith, C.~Chiang, M.~Sanjabi, and A.~Talwalkar, ``Federated multi-task
  learning,'' in \emph{{NeurIPS}}, 2017.

\bibitem[Bonawitz et~al.(2017{\natexlab{b}})Bonawitz, Ivanov, Kreuter,
  Marcedone, McMahan, Patel, Ramage, Segal, and
  Seth]{DBLP:conf/ccs/BonawitzIKMMPRS17}
K.~A. Bonawitz, V.~Ivanov, B.~Kreuter, A.~Marcedone, H.~B. McMahan, S.~Patel,
  D.~Ramage, A.~Segal, and K.~Seth, ``Practical secure aggregation for
  privacy-preserving machine learning,'' in \emph{{CCS}}, 2017.

\bibitem[Mohri et~al.(2019)Mohri, Sivek, and Suresh]{mohri2019agnostic}
M.~Mohri, G.~Sivek, and A.~T. Suresh, ``Agnostic federated learning,'' in
  \emph{{ICML}}, 2019.

\bibitem[Yurochkin et~al.(2019)Yurochkin, Agarwal, Ghosh, Greenewald, Hoang,
  and Khazaeni]{DBLP:conf/icml/YurochkinAGGHK19}
M.~Yurochkin, M.~Agarwal, S.~Ghosh, K.~H. Greenewald, T.~N. Hoang, and
  Y.~Khazaeni, ``Bayesian nonparametric federated learning of neural
  networks,'' in \emph{{ICML}}, 2019.

\bibitem[Peng et~al.(2020)Peng, Huang, Zhu, and
  Saenko]{DBLP:conf/iclr/PengHZS20}
X.~Peng, Z.~Huang, Y.~Zhu, and K.~Saenko, ``Federated adversarial domain
  adaptation,'' in \emph{{ICLR}}, 2020.

\bibitem[Wang et~al.(2020{\natexlab{a}})Wang, Yurochkin, Sun, Papailiopoulos,
  and Khazaeni]{wang2020federated}
H.~Wang, M.~Yurochkin, Y.~Sun, D.~S. Papailiopoulos, and Y.~Khazaeni,
  ``Federated learning with matched averaging,'' in \emph{{ICLR}}, 2020.

\bibitem[Hamer et~al.(2020)Hamer, Mohri, and Suresh]{DBLP:conf/icml/HamerMS20}
J.~Hamer, M.~Mohri, and A.~T. Suresh, ``Fedboost: {A} communication-efficient
  algorithm for federated learning,'' in \emph{{ICML}}, 2020.

\bibitem[Yu et~al.(2020)Yu, Rawat, Menon, and Kumar]{DBLP:conf/icml/YuRMK20}
F.~X. Yu, A.~S. Rawat, A.~K. Menon, and S.~Kumar, ``Federated learning with
  only positive labels,'' in \emph{{ICML}}, 2020.

\bibitem[Rothchild et~al.(2020)Rothchild, Panda, Ullah, Ivkin, Stoica,
  Braverman, Gonzalez, and Arora]{DBLP:conf/icml/RothchildPUISB020}
D.~Rothchild, A.~Panda, E.~Ullah, N.~Ivkin, I.~Stoica, V.~Braverman,
  J.~Gonzalez, and R.~Arora, ``Fetchsgd: Communication-efficient federated
  learning with sketching,'' in \emph{{ICML}}, 2020.

\bibitem[Karimireddy et~al.(2020)Karimireddy, Kale, Mohri, Reddi, Stich, and
  Suresh]{DBLP:conf/icml/KarimireddyKMRS20}
S.~P. Karimireddy, S.~Kale, M.~Mohri, S.~J. Reddi, S.~U. Stich, and A.~T.
  Suresh, ``{SCAFFOLD:} stochastic controlled averaging for federated
  learning,'' in \emph{{ICML}}, 2020.

\bibitem[Malinovskiy et~al.(2020)Malinovskiy, Kovalev, Gasanov, Condat, and
  Richt{\'{a}}rik]{DBLP:conf/icml/MalinovskiyKGCR20}
G.~Malinovskiy, D.~Kovalev, E.~Gasanov, L.~Condat, and P.~Richt{\'{a}}rik,
  ``From local {SGD} to local fixed-point methods for federated learning,'' in
  \emph{{ICML}}, 2020.

\bibitem[Fallah et~al.(2020)Fallah, Mokhtari, and
  Ozdaglar]{DBLP:conf/nips/0001MO20}
A.~Fallah, A.~Mokhtari, and A.~E. Ozdaglar, ``Personalized federated learning
  with theoretical guarantees: {A} model-agnostic meta-learning approach,'' in
  \emph{NeurIPS}, 2020.

\bibitem[Dinh et~al.(2020)Dinh, Tran, and Nguyen]{DBLP:conf/nips/DinhTN20}
C.~T. Dinh, N.~H. Tran, and T.~D. Nguyen, ``Personalized federated learning
  with moreau envelopes,'' in \emph{NeurIPS}, 2020.

\bibitem[Pathak and Wainwright(2020)]{DBLP:conf/nips/PathakW20}
R.~Pathak and M.~J. Wainwright, ``Fedsplit: an algorithmic framework for fast
  federated optimization,'' in \emph{NeurIPS}, 2020.

\bibitem[Marfoq et~al.(2020)Marfoq, Xu, Neglia, and
  Vidal]{DBLP:conf/nips/MarfoqXNV20}
O.~Marfoq, C.~Xu, G.~Neglia, and R.~Vidal, ``Throughput-optimal topology design
  for cross-silo federated learning,'' in \emph{NeurIPS}, 2020.

\bibitem[Wang et~al.(2020{\natexlab{b}})Wang, Liu, Liang, Joshi, and
  Poor]{DBLP:conf/nips/WangLLJP20}
J.~Wang, Q.~Liu, H.~Liang, G.~Joshi, and H.~V. Poor, ``Tackling the objective
  inconsistency problem in heterogeneous federated optimization,'' in
  \emph{NeurIPS}, 2020.

\bibitem[Reisizadeh et~al.(2020)Reisizadeh, Farnia, Pedarsani, and
  Jadbabaie]{DBLP:conf/nips/ReisizadehFPJ20}
A.~Reisizadeh, F.~Farnia, R.~Pedarsani, and A.~Jadbabaie, ``Robust federated
  learning: The case of affine distribution shifts,'' in \emph{NeurIPS}, 2020.

\bibitem[He et~al.(2020{\natexlab{a}})He, Annavaram, and
  Avestimehr]{DBLP:conf/nips/0001AA20}
C.~He, M.~Annavaram, and S.~Avestimehr, ``Group knowledge transfer: Federated
  learning of large cnns at the edge,'' in \emph{NeurIPS}, 2020.

\bibitem[Grammenos et~al.(2020)Grammenos, Mendoza{-}Smith, Crowcroft, and
  Mascolo]{DBLP:conf/nips/GrammenosMCM20}
A.~Grammenos, R.~Mendoza{-}Smith, J.~Crowcroft, and C.~Mascolo, ``Federated
  principal component analysis,'' in \emph{NeurIPS}, 2020.

\bibitem[Dai et~al.(2020)Dai, Low, and Jaillet]{DBLP:conf/nips/DaiLJ20}
Z.~Dai, B.~K.~H. Low, and P.~Jaillet, ``Federated bayesian optimization via
  thompson sampling,'' in \emph{NeurIPS}, 2020.

\bibitem[Yuan and Ma(2020)]{DBLP:conf/nips/YuanM20}
H.~Yuan and T.~Ma, ``Federated accelerated stochastic gradient descent,'' in
  \emph{NeurIPS}, 2020.

\bibitem[Ghosh et~al.(2020)Ghosh, Chung, Yin, and
  Ramchandran]{DBLP:conf/nips/GhoshCYR20}
A.~Ghosh, J.~Chung, D.~Yin, and K.~Ramchandran, ``An efficient framework for
  clustered federated learning,'' in \emph{NeurIPS}, 2020.

\bibitem[Dubey and Pentland(2020)]{DBLP:conf/nips/DubeyP20}
A.~Dubey and A.~S. Pentland, ``Differentially-private federated linear
  bandits,'' in \emph{NeurIPS}, 2020.

\bibitem[Deng et~al.(2020)Deng, Kamani, and Mahdavi]{DBLP:conf/nips/DengKM20}
Y.~Deng, M.~M. Kamani, and M.~Mahdavi, ``Distributionally robust federated
  averaging,'' in \emph{NeurIPS}, 2020.

\bibitem[Lin et~al.(2020)Lin, Kong, Stich, and Jaggi]{DBLP:conf/nips/LinKSJ20}
T.~Lin, L.~Kong, S.~U. Stich, and M.~Jaggi, ``Ensemble distillation for robust
  model fusion in federated learning,'' in \emph{NeurIPS}, 2020.

\bibitem[Niu et~al.(2020)Niu, Wu, Tang, Hua, Jia, Lv, Wu, and
  Chen]{DBLP:conf/mobicom/NiuWTHJLWC20}
C.~Niu, F.~Wu, S.~Tang, L.~Hua, R.~Jia, C.~Lv, Z.~Wu, and G.~Chen,
  ``Billion-scale federated learning on mobile clients: a submodel design with
  tunable privacy,'' in \emph{MobiCom}, 2020.

\bibitem[Gu et~al.(2020)Gu, Dang, Li, and Huang]{DBLP:conf/kdd/GuDLH20}
B.~Gu, Z.~Dang, X.~Li, and H.~Huang, ``Federated doubly stochastic kernel
  learning for vertically partitioned data,'' in \emph{{KDD}}, 2020.

\bibitem[Zhang et~al.(2021)Zhang, Sapra, Fidler, Yeung, and
  Alvarez]{DBLP:conf/iclr/ZhangSFYA21}
M.~Zhang, K.~Sapra, S.~Fidler, S.~Yeung, and J.~M. Alvarez, ``Personalized
  federated learning with first order model optimization,'' in \emph{{ICLR}},
  2021.

\bibitem[Chen and Chao(2021)]{DBLP:conf/iclr/ChenC21}
H.~Chen and W.~Chao, ``Fedbe: Making bayesian model ensemble applicable to
  federated learning,'' in \emph{{ICLR}}, 2021.

\bibitem[Li et~al.(2021)Li, Jiang, Zhang, Kamp, and
  Dou]{DBLP:conf/iclr/LiJZKD21}
X.~Li, M.~Jiang, X.~Zhang, M.~Kamp, and Q.~Dou, ``Fedbn: Federated learning on
  non-iid features via local batch normalization,'' in \emph{{ICLR}}, 2021.

\bibitem[Acar et~al.(2021{\natexlab{b}})Acar, Zhao, Navarro, Mattina,
  Whatmough, and Saligrama]{DBLP:conf/iclr/AcarZNMWS21}
D.~A.~E. Acar, Y.~Zhao, R.~M. Navarro, M.~Mattina, P.~N. Whatmough, and
  V.~Saligrama, ``Federated learning based on dynamic regularization,'' in
  \emph{{ICLR}}, 2021.

\bibitem[Al{-}Shedivat et~al.(2021)Al{-}Shedivat, Gillenwater, Xing, and
  Rostamizadeh]{DBLP:conf/iclr/Al-ShedivatGXR21}
M.~Al{-}Shedivat, J.~Gillenwater, E.~P. Xing, and A.~Rostamizadeh, ``Federated
  learning via posterior averaging: {A} new perspective and practical
  algorithms,'' in \emph{{ICLR}}, 2021.

\bibitem[Jeong et~al.(2021)Jeong, Yoon, Yang, and
  Hwang]{DBLP:conf/iclr/JeongYYH21}
W.~Jeong, J.~Yoon, E.~Yang, and S.~J. Hwang, ``Federated semi-supervised
  learning with inter-client consistency {\&} disjoint learning,'' in
  \emph{{ICLR}}, 2021.

\bibitem[Yoon et~al.(2021)Yoon, Shin, Hwang, and
  Yang]{DBLP:conf/iclr/YoonSHY21}
T.~Yoon, S.~Shin, S.~J. Hwang, and E.~Yang, ``Fedmix: Approximation of mixup
  under mean augmented federated learning,'' in \emph{{ICLR}}, 2021.

\bibitem[Murata and Suzuki(2021)]{DBLP:conf/icml/MurataS21}
T.~Murata and T.~Suzuki, ``Bias-variance reduced local {SGD} for less
  heterogeneous federated learning,'' in \emph{{ICML}}, 2021.

\bibitem[Fraboni et~al.(2021)Fraboni, Vidal, Kameni, and
  Lorenzi]{DBLP:conf/icml/FraboniVKL21}
Y.~Fraboni, R.~Vidal, L.~Kameni, and M.~Lorenzi, ``Clustered sampling:
  Low-variance and improved representativity for clients selection in federated
  learning,'' in \emph{{ICML}}, 2021.

\bibitem[Kairouz et~al.(2021)Kairouz, Liu, and
  Steinke]{DBLP:conf/icml/KairouzL021}
P.~Kairouz, Z.~Liu, and T.~Steinke, ``The distributed discrete gaussian
  mechanism for federated learning with secure aggregation,'' in \emph{{ICML}},
  2021.

\bibitem[Huang et~al.(2021)Huang, Li, Song, and
  Yang]{DBLP:conf/icml/HuangL0021}
B.~Huang, X.~Li, Z.~Song, and X.~Yang, ``{FL-NTK:} {A} neural tangent
  kernel-based framework for federated learning analysis,'' in \emph{{ICML}},
  2021.

\bibitem[Yuan et~al.(2021)Yuan, Guo, Xu, Ying, and
  Yang]{DBLP:conf/icml/YuanGXYY21}
Z.~Yuan, Z.~Guo, Y.~Xu, Y.~Ying, and T.~Yang, ``Federated deep {AUC}
  maximization for hetergeneous data with a constant communication
  complexity,'' in \emph{{ICML}}, 2021.

\bibitem[Lai et~al.(2021{\natexlab{b}})Lai, Zhu, Madhyastha, and
  Chowdhury]{DBLP:conf/osdi/LaiZMC21}
F.~Lai, X.~Zhu, H.~V. Madhyastha, and M.~Chowdhury, ``Oort: Efficient federated
  learning via guided participant selection,'' in \emph{{OSDI}}, 2021.

\bibitem[Sav et~al.(2021{\natexlab{b}})Sav, Pyrgelis, Troncoso{-}Pastoriza,
  Froelicher, Bossuat, Sousa, and Hubaux]{DBLP:conf/ndss/SavPTFBSH21}
S.~Sav, A.~Pyrgelis, J.~R. Troncoso{-}Pastoriza, D.~Froelicher, J.~Bossuat,
  J.~S. Sousa, and J.~Hubaux, ``{POSEIDON:} privacy-preserving federated neural
  network learning,'' in \emph{{NDSS}}, 2021.

\bibitem[LeCun et~al.(1998)LeCun, Bottou, Bengio, and
  Haffner]{lecun1998gradient}
Y.~LeCun, L.~Bottou, Y.~Bengio, and P.~Haffner, ``Gradient-based learning
  applied to document recognition,'' \emph{Proceedings of the IEEE}, 1998.

\bibitem[Go et~al.(2009)Go, Bhayani, and Huang]{go2009twitter}
A.~Go, R.~Bhayani, and L.~Huang, ``Twitter sentiment classification using
  distant supervision,'' \emph{CS224N project report, Stanford}, 2009.

\bibitem[Caldas et~al.(2018{\natexlab{b}})Caldas, Duddu, Wu, Li,
  Kone{\v{c}}n{\`y}, McMahan, Smith, and Talwalkar]{LEAFData}
S.~Caldas, S.~M.~K. Duddu, P.~Wu, T.~Li, J.~Kone{\v{c}}n{\`y}, H.~B. McMahan,
  V.~Smith, and A.~Talwalkar, ``Leaf: A benchmark for federated settings,''
  \emph{arXiv}, 2018.

\bibitem[Beutel et~al.(2020)Beutel, Topal, Mathur, Qiu, Parcollet, and
  Lane]{flower}
D.~J. Beutel, T.~Topal, A.~Mathur, X.~Qiu, T.~Parcollet, and N.~D. Lane,
  ``Flower: A friendly federated learning research framework,'' \emph{arXiv},
  2020.

\bibitem[Chai et~al.(2022)Chai, Wang, Zhang, Yang, Cai, Chen, and Yang]{FedSVD}
D.~Chai, L.~Wang, J.~Zhang, L.~Yang, S.~Cai, K.~Chen, and Q.~Yang, ``Practical
  lossless federated singular vector decomposition over billion-scale data,''
  in \emph{{KDD}}, 2022.

\bibitem[He et~al.(2020{\natexlab{b}})He, Li, So, Zhang, Wang, Wang, Vepakomma,
  Singh, Qiu, Shen, et~al.]{he2020fedml}
C.~He, S.~Li, J.~So, M.~Zhang, H.~Wang, X.~Wang, P.~Vepakomma, A.~Singh,
  H.~Qiu, L.~Shen \emph{et~al.}, ``Fedml: A research library and benchmark for
  federated machine learning,'' \emph{arXiv}, 2020.

\bibitem[Bagdasaryan et~al.(2020)Bagdasaryan, Veit, Hua, Estrin, and
  Shmatikov]{bagdasaryan2018backdoor}
E.~Bagdasaryan, A.~Veit, Y.~Hua, D.~Estrin, and V.~Shmatikov, ``How to backdoor
  federated learning,'' in \emph{{AISTATS}}, 2020.

\bibitem[Hidano et~al.(2018)Hidano, Murakami, Katsumata, Kiyomoto, and
  Hanaoka]{hidano2018model}
S.~Hidano, T.~Murakami, S.~Katsumata, S.~Kiyomoto, and G.~Hanaoka, ``Model
  inversion attacks for online prediction systems: Without knowledge of
  non-sensitive attributes,'' \emph{IEICE Transactions on Information and
  Systems}, 2018.

\bibitem[Fredrikson et~al.(2015)Fredrikson, Jha, and
  Ristenpart]{fredrikson2015model}
M.~Fredrikson, S.~Jha, and T.~Ristenpart, ``Model inversion attacks that
  exploit confidence information and basic countermeasures,'' in
  \emph{{SIGSAC}}, 2015.

\bibitem[Zhang et~al.(2020)Zhang, Zhang, Chen, and Yu]{zhang2020gan}
J.~Zhang, J.~Zhang, J.~Chen, and S.~Yu, ``Gan enhanced membership inference: A
  passive local attack in federated learning,'' in \emph{{ICC}}, 2020.

\end{thebibliography}
%
%
%

%

\begin{IEEEbiography}[{\includegraphics[width=1in,height=1.25in,clip,keepaspectratio]{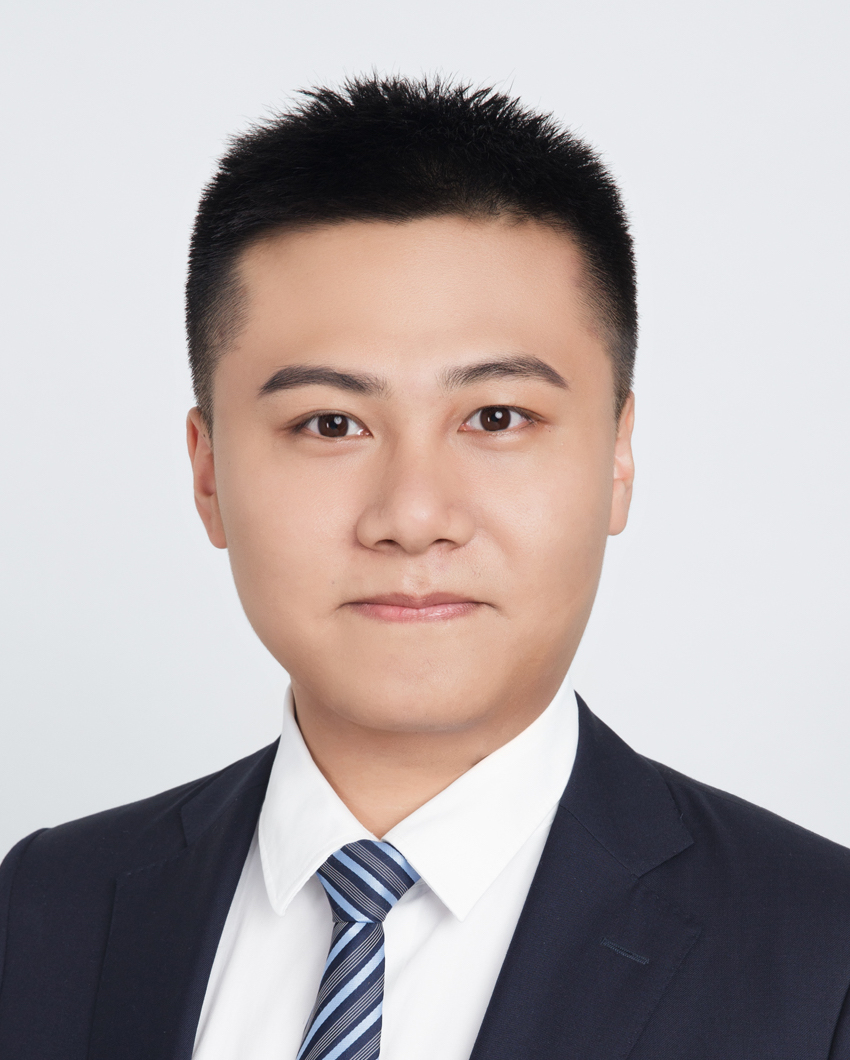}}]{Di Chai}
is a Ph.D. student in computer science and engineering at Hong Kong University of Science and Technology (HKUST). He got his master degree of science from HKUST in 2018. His research interests include federated learning and privacy-preserving machine learning.
\end{IEEEbiography}	

\vspace{-12mm}

\begin{IEEEbiography}[{\includegraphics[width=1in,height=1.25in,clip,keepaspectratio]{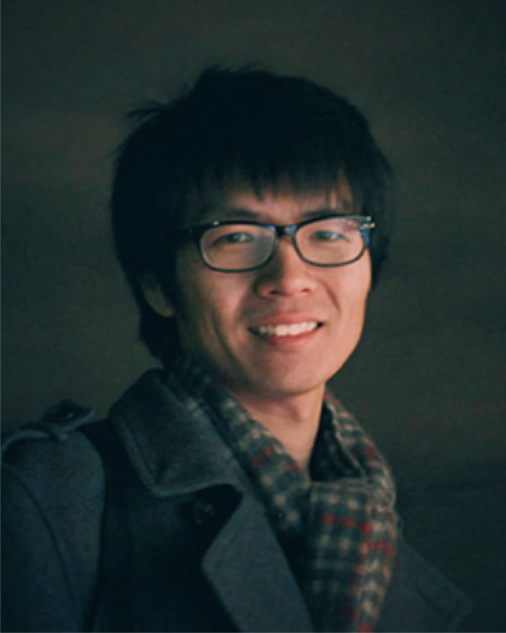}}]{Leye Wang}
	is an assistant professor at Key Lab of High Confidence Software Technolo- gies (Peking University), MOE, and School of Computer Science, Peking University, China. He received a Ph.D. in computer science from TELECOM SudParis and University Paris 6, France, in 2016. He was a postdoc researcher with Hong Kong University of Science and Technology. His research interests include ubiquitous computing, mobile crowdsensing, and urban computing.
\end{IEEEbiography}	

\vspace{-12mm}

\begin{IEEEbiography}[{\includegraphics[width=1in,height=1.25in,clip,keepaspectratio]{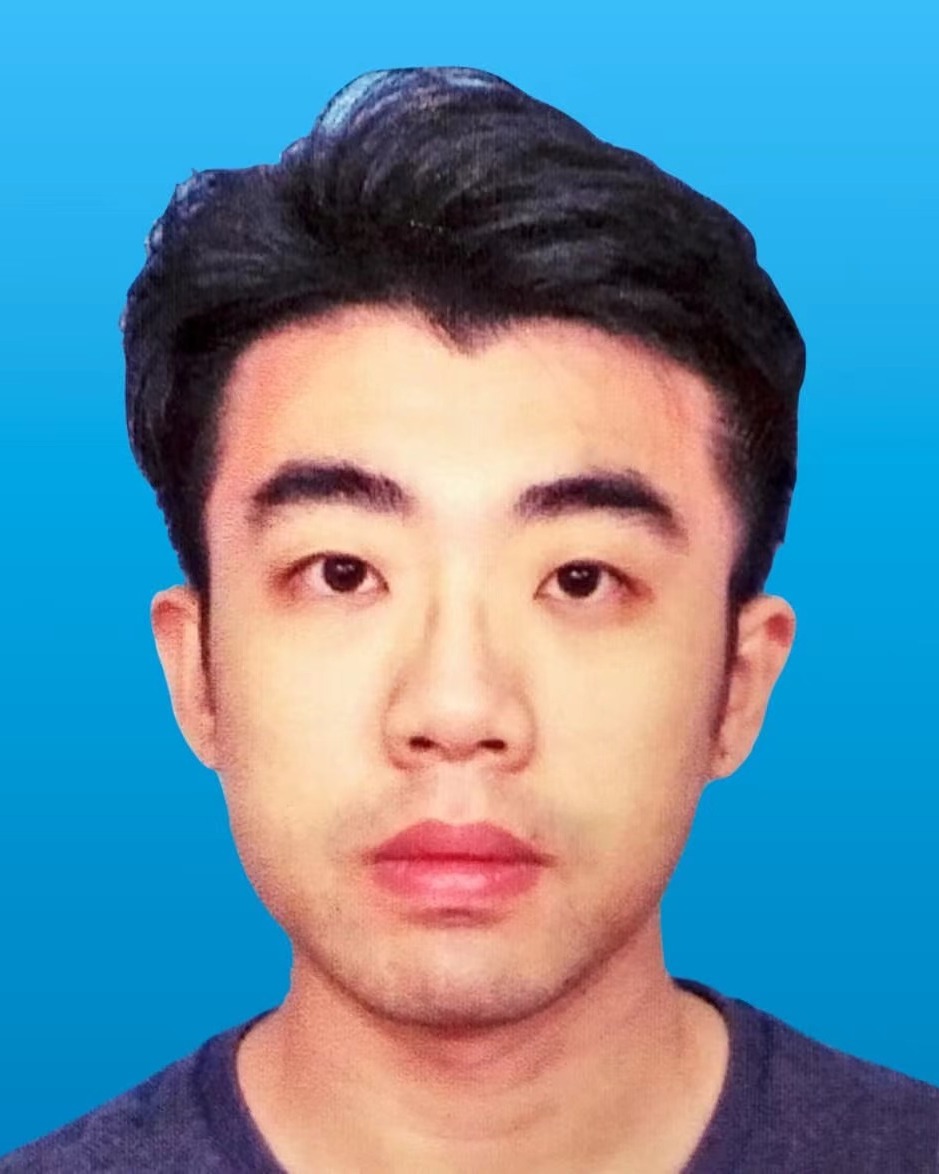}}]{Liu Yang}
	is a PhD student of computer science at iSINGLab, Hong Kong University of Science and Technology (HKUST). He is under supervision of Prof. Qiang Yang and Prof. Kai Chen. His research interests include federated learning and recommendation system. Before pursuing PhD, he received his BEng and MSc from Sun Yat-sen University and HKUST, respectively.
\end{IEEEbiography}	

\vspace{-12mm}

\begin{IEEEbiography}[{\includegraphics[width=1in,height=1.25in,clip,keepaspectratio]{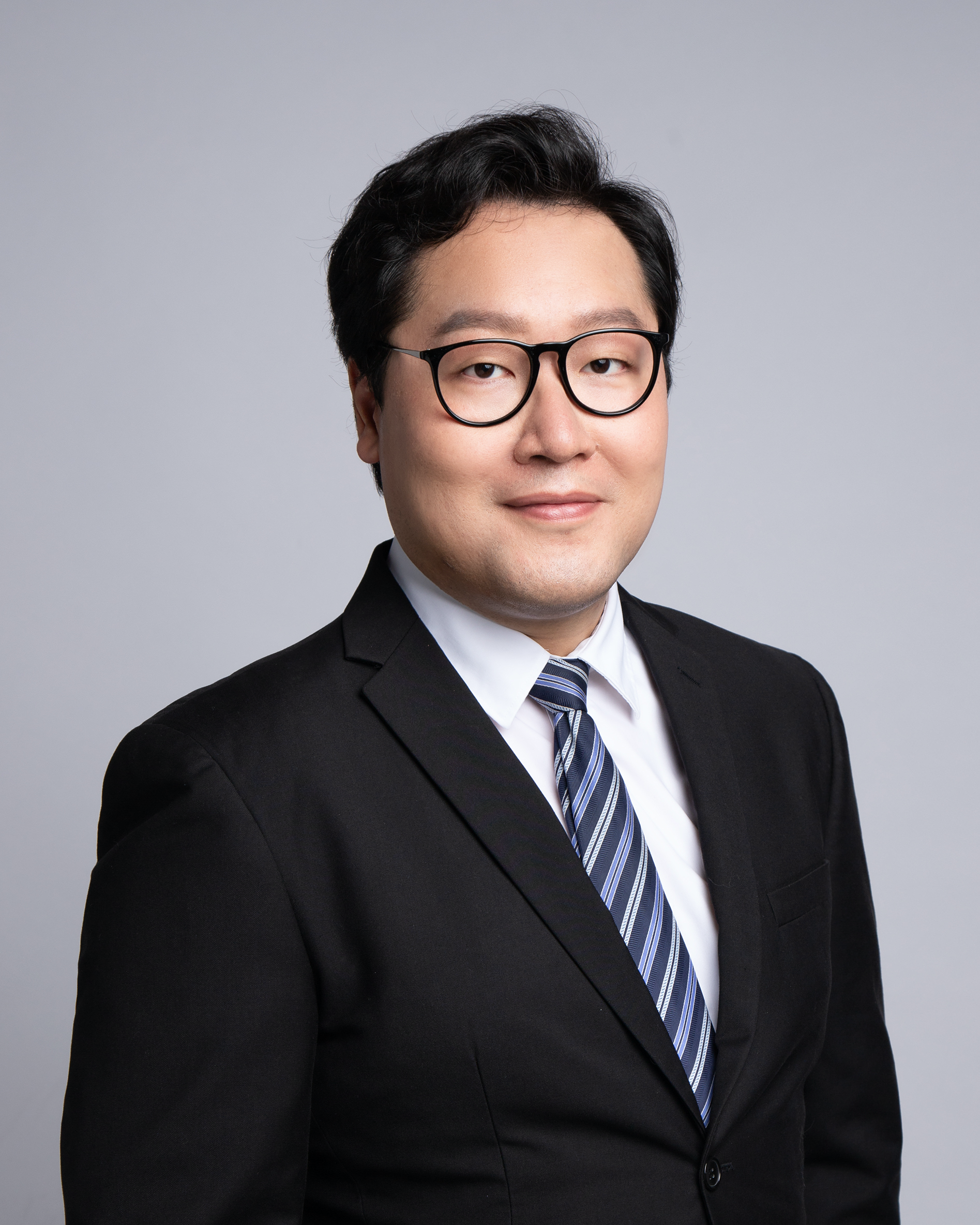}}]{Junxue Zhang}
	is a Ph.D. candidate of computer science at iSINGLab, Hong Kong Univeristy of Science and Technology. He is under supervision of Prof. Kai CHEN. His reseach interests are data center networking, AI systems and privacy preserving computation. He has published papers in SIGCOMM/CoNEXT.
\end{IEEEbiography}	

\vspace{-12mm}

\begin{IEEEbiography}[{\includegraphics[width=1in,height=1.25in,clip,keepaspectratio]{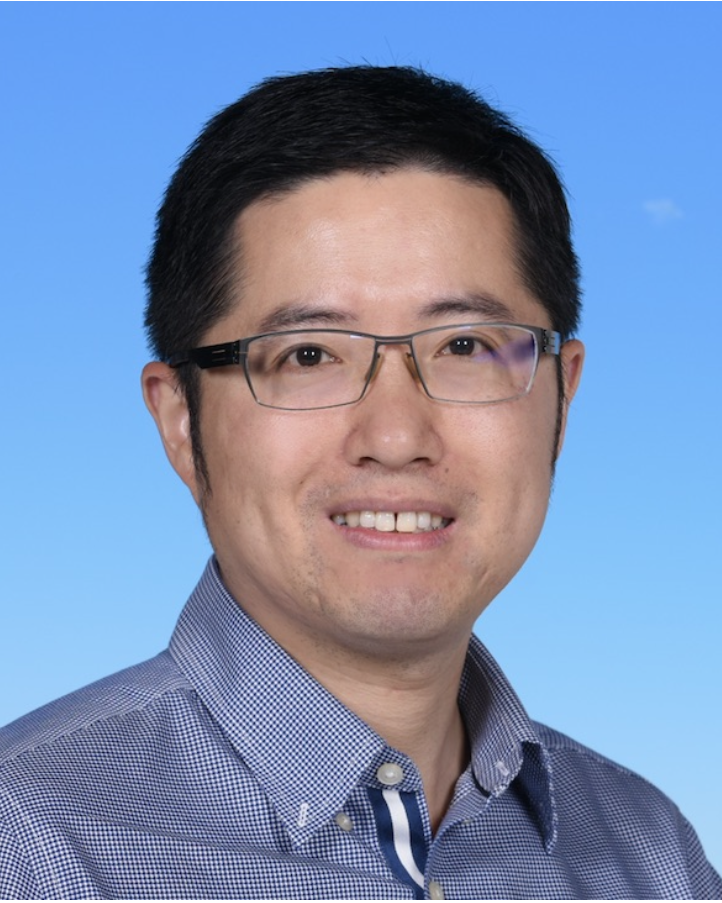}}]{Kai Chen}
	is the Professor with the Department of Computer Science and Engineering, Hong Kong University of Science and Technology, Hong Kong. He received his Ph.D. degree in Computer Science from Northwestern University, Evanston, IL, USA in 2012. His research interests include data center networking, machine learning systems and privacy-preserving computing.
\end{IEEEbiography}

\vspace{-12mm}

\begin{IEEEbiography}[{\includegraphics[width=1in,height=1.25in,clip,keepaspectratio]{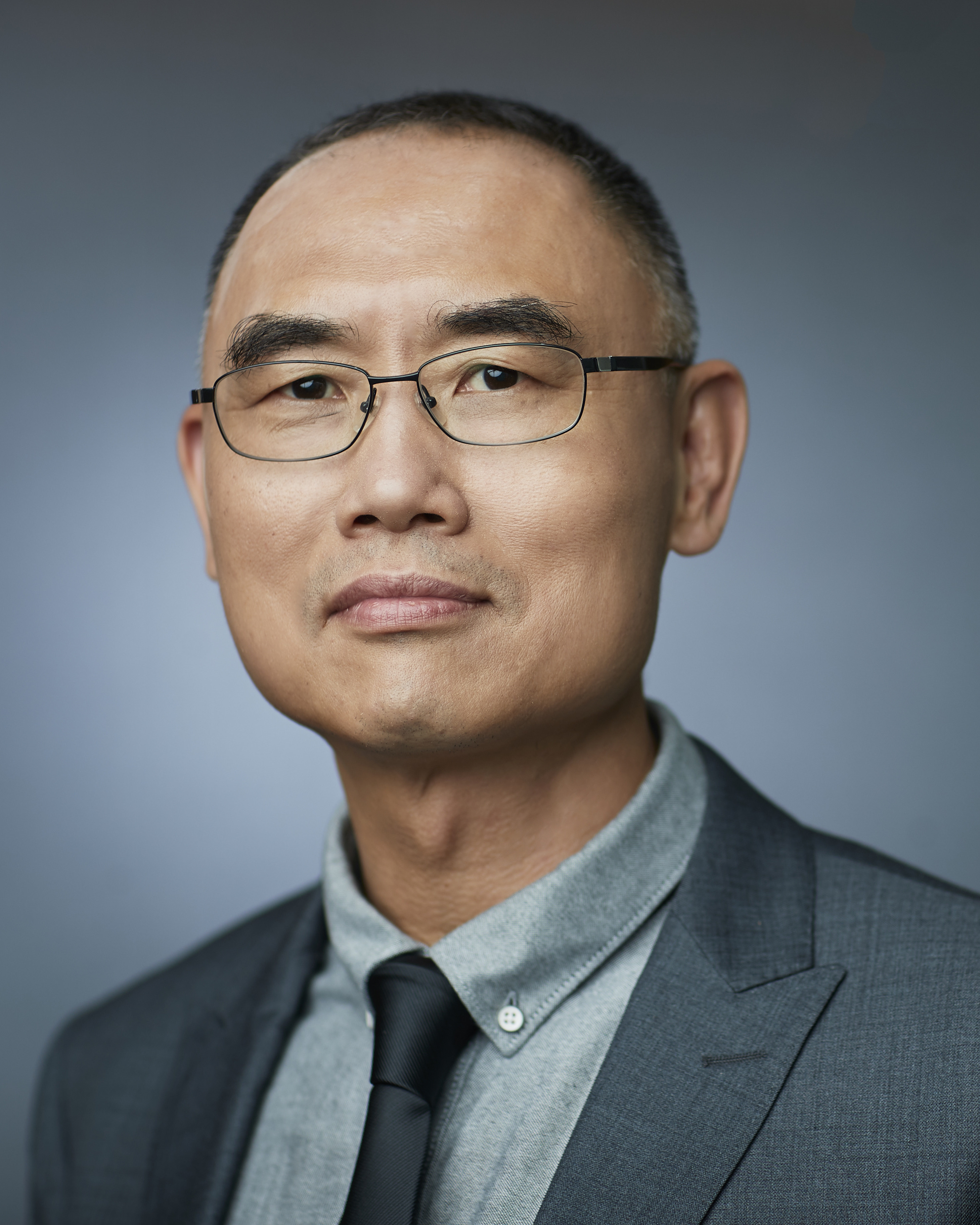}}]{Qiang Yang}
	is a Fellow of Canadian Academy of Engineering (CAE) and Royal Society of Canada (RSC), Chief Artificial Intelligence Officer of WeBank, a Chair Professor of Computer Science and Engineering Department at Hong Kong University of Science and Technology (HKUST). He is the Conference Chair of AAAI-21, the Honorary Vice President of Chinese Association for Artificial Intelligence(CAAI) , the President of Hong Kong Society of Artificial Intelligence and Robotics(HKSAIR) and the President of Investment Technology League (ITL). He is a fellow of AAAI, ACM, CAAI, IEEE, IAPR, AAAS. He was the Founding Editor in Chief of the ACM Transactions on Intelligent Systems and Technology (ACM TIST) and the Founding Editor in Chief of IEEE Transactions on Big Data (IEEE TBD). He received the ACM SIGKDD Distinguished Service Award in 2017. He had been the Founding Director of the Huawei’s Noah’s Ark Research Lab between 2012 and 2015, the Founding Director of HKUST’s Big Data Institute, the Founder of 4Paradigm and the President of IJCAI (2017-2019). His research interests are artificial intelligence, machine learning, data mining and planning.
\end{IEEEbiography}









\end{document}